%% file: main.tex
\documentclass[journal]{IEEEtran}

\usepackage{pifont}

\usepackage{bbm}
\usepackage{moreverb,url}
\usepackage{algorithm}
\usepackage{algpseudocode}
\usepackage{multirow}
\usepackage{graphicx}
\usepackage{booktabs}
\usepackage{bbm}
\usepackage{soul}
\usepackage{amssymb}
\usepackage{cite}
\usepackage[colorlinks,bookmarksopen,bookmarksnumbered,citecolor=red,urlcolor=red]{hyperref}

\input{sections/notation}

\usepackage[capitalise,nameinlink]{cleveref} 

\begin{document}

\title{Versatile Behavior Diffusion for \\ Generalized Traffic Agent Simulation}

\author{Zhiyu Huang,~\IEEEmembership{Member,~IEEE,}
        Zixu Zhang, Ameya Vaidya, Yuxiao Chen,\\
        Jaime F. Fisac, and Chen Lv,~\IEEEmembership{Senior Member,~IEEE}%
\thanks{Zhiyu Huang and Chen Lv are with the School of Mechanical and Aerospace Engineering, Nanyang Technological University, Singapore (E-mail: \texttt{\small zhiyu001@e.ntu.edu.sg; lyuchen@ntu.edu.sg}).}%
\thanks{Zixu Zhang, Ameya Vaidya, and Jaime F. Fisac are with the Department of Electrical and Computer Engineering, Princeton University, Princeton, NJ, USA (E-mail: \texttt{\small \{zixuz, avaidya, jfisac\}@princeton.edu}).}%
\thanks{Yuxiao Chen is with NVIDIA Research, Santa Clara, CA, USA (E-mail: \texttt{\small yuxiaoc@nvidia.com}).}%
\thanks{Zhiyu Huang and Zixu Zhang contributed equally to this work.}%
}




\maketitle

\begin{abstract}
\input{sections/abstract}
\end{abstract}


\input{sections/intro}

\input{sections/related_works}
\input{sections/method/problem_formulation}

\input{sections/method/model}

\input{sections/experiment}

\input{sections/conclusions}

\bibliographystyle{IEEEtran}
\bibliography{reference} 

\end{document}

%% file: sections/notation.tex
\usepackage{amsmath}
\usepackage{bm}

\usepackage{ifthen}
\newboolean{include-notes}
\newboolean{include-new}
\newboolean{include-remove}
\setboolean{include-notes}{true}
\setboolean{include-new}{true}
\setboolean{include-remove}{false}

\usepackage{xcolor}
\usepackage[normalem]{ulem}
\definecolor{newcolor}{rgb}{0, 0, 1}
\definecolor{removecolor}{rgb}{1, 0, 0}
\newcommand{\jaime}[1]{\ifthenelse{\boolean{include-notes}}{\textcolor{orange}{\textbf{JFF:} #1}}{}}
\newcommand{\zzx}[1]{\ifthenelse{\boolean{include-notes}}{\textcolor{teal}{\textbf{Z:} #1}}{}}
\newcommand{\zhiyu}[1]{\ifthenelse{\boolean{include-notes}}{\textcolor{Peach}{\textbf{H:} #1}}{}}
\newcommand{\ameya}[1]{\ifthenelse{\boolean{include-notes}}{\textcolor{violet}{\textbf{A:} #1}}{}}
\newcommand{\remove}[1]{\ifthenelse{\boolean{include-remove}}{\textcolor{removecolor}{\sout{#1}}}{}}
\newcommand{\new}[1]{\ifthenelse{\boolean{include-new}}{\textcolor{newcolor}{#1}}{#1}}

\definecolor{porange}{HTML}{E77500} 

\usepackage{lipsum}

\newcommand{\model}{{Versatile Behavior Diffusion}}
\newcommand{\modelabb}{{VBD}}

\newcommand{\denoiser}{\mathcal{D}_\theta}
\newcommand{\predictor}{\mathcal{P}_\psi}
\newcommand{\reals}{\mathbb{R}}

\newcommand{\dataset}{\mathbb{D}}
\newcommand{\distrdata}{p}
\newcommand{\distr}{p}

\newcommand{\uniform}{{\mathcal{U}}}
\newcommand{\categorical}{\text{Cat}}

\newcommand{\expectation}{{\mathbb{E}}}

\usepackage{cancel}

\DeclareMathOperator*{\argmax}{arg\,max}

\newcommand{\nagents}{{A}}

\newcommand{\scenario}{\mathcal{S}}
\newcommand{\state}{{x}}
\newcommand{\traj}{{\mathbf{x}}}

\newcommand{\ctrlseq}{\mathbf{u}}

\newcommand{\score}{\mathbf{s}}

\newcommand{\condition}{\mathbf{c}}

\newcommand{\nx}{{D_x}}
\newcommand{\nc}{{D_u}}

\newcommand{\dyn}{{f}}

\newcommand{\tdisc}{{t}}
\newcommand{\thorizon}{{T}}

\newcommand{\cost}{\mathcal{J}}
\newcommand{\hparam}{{\theta}}

%% file: sections/abstract.tex
Existing traffic simulation models often fall short in capturing the intricacies of real-world scenarios, particularly the interactive behaviors among multiple traffic participants, thereby limiting their utility in the evaluation and validation of autonomous driving systems. We introduce Versatile Behavior Diffusion (VBD), a novel traffic scenario generation framework based on diffusion generative models that synthesizes scene-consistent, realistic, and controllable multi-agent interactions. VBD achieves strong performance in closed-loop traffic simulation, generating scene-consistent agent behaviors that reflect complex agent interactions. A key capability of VBD is inference-time scenario editing through multi-step refinement, guided by behavior priors and model-based optimization objectives, enabling flexible and controllable behavior generation. Despite being trained on real-world traffic datasets with only normal conditions, we introduce conflict-prior and game-theoretic guidance approaches. These approaches enable the generation of interactive, customizable, or long-tail safety-critical scenarios, which are essential for comprehensive testing and validation of autonomous driving systems. Extensive experiments validate the effectiveness and versatility of VBD and highlight its promise as a foundational tool for advancing traffic simulation and autonomous vehicle development. Project website: \href{https://sites.google.com/view/versatile-behavior-diffusion}{https://sites.google.com/view/versatile-behavior-diffusion}.

%% file: sections/intro.tex
\section{Introduction}
\label{sec:intro}

\IEEEPARstart{S}{imulation} testing is crucial for ensuring the safe deployment of autonomous driving systems. Traditionally, these assessments rely on replaying logged trajectories from driving datasets in simulation environments \cite{li2022metadrive}. However, because other traffic participants in the replay do not react to the autonomous vehicle's actions, log-replay methods often fail to account for interactive traffic scenarios, such as pedestrians stepping into crosswalks or vehicles negotiating right-of-way at four-way stops. This limitation introduces a significant simulation-to-reality gap, where simulated test results may not translate to real-world performance, undermining the credibility of simulation-based testing. 

\begin{figure}[htp]
    \centering
    \includegraphics[width=\linewidth]{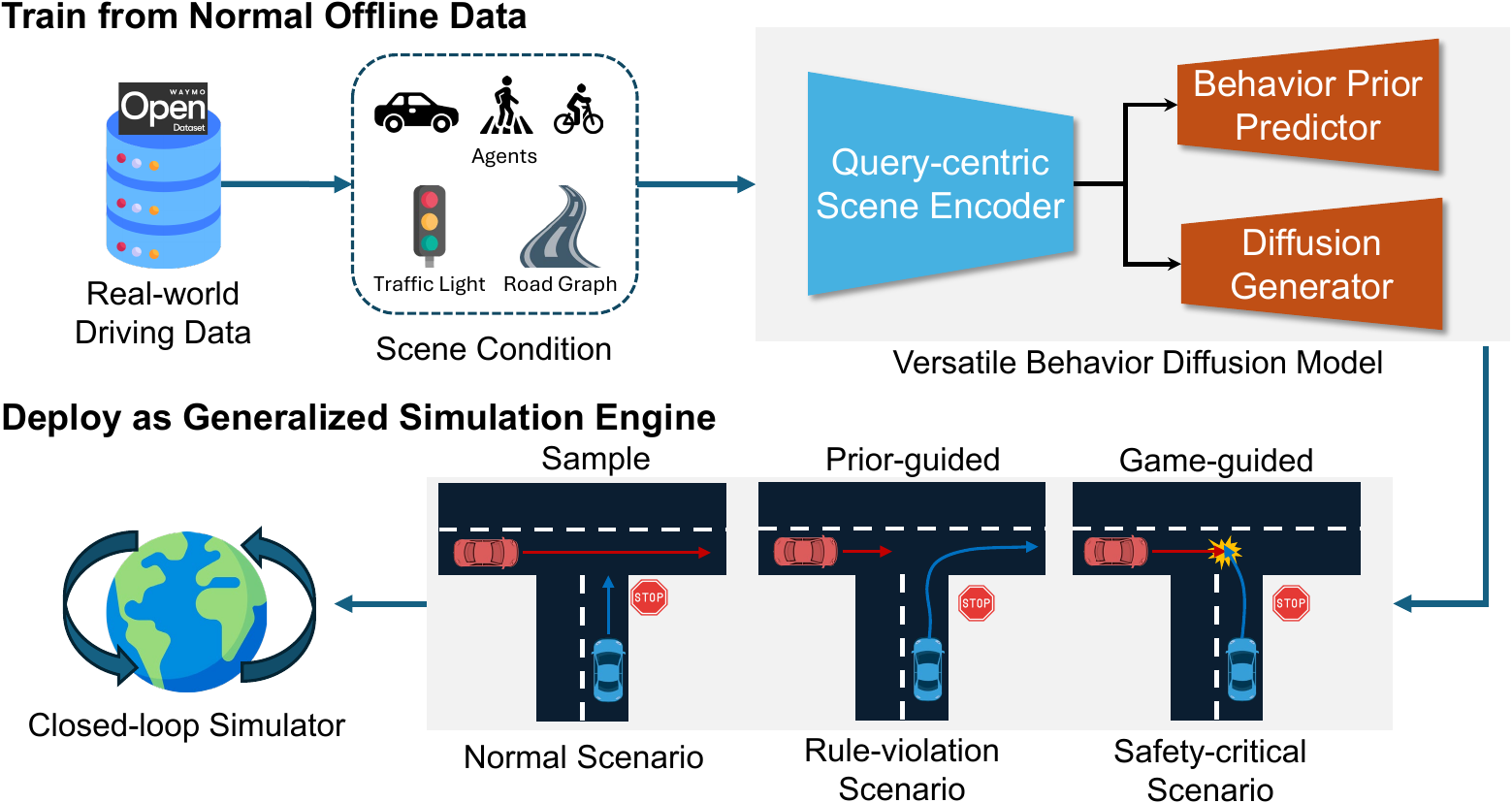}
    \caption{Illustration of the VBD framework. The model consists of a query-centric Transformer scene condition encoder, a marginal multi-modal behavior predictor for individual agents, and a diffusion generator for joint multi-agent behavior. In training, the model only utilizes commonly used traffic datasets. In testing, the model's versatility is showcased through various purposes enabled by simple sampling and different guidance structures (e.g., prior guidance for sophisticated target agents' behavior control and game-theoretic cost guidance for adversarial behaviors). VBD functions as a generalized simulation engine and operates effectively in a closed-loop traffic simulator.}
    \label{fig:0}
\end{figure}

To address this challenge, traffic simulators that incorporate reactive traffic behavior models have become increasingly popular. These models account for interactions between agents and their surroundings, enabling the synthesis of scenarios that better reflect the complex dynamics of real-world traffic. However, conventional heuristic-based traffic modeling approaches \cite{treiber2000congested, kesting2007general} often fail to scale to the complexity of realistic traffic maneuvers. As a result, recent efforts \cite{shi2022motion, igl2022symphony, nayakanti2023wayformer, zhang2023trafficbots, feng2023trafficgen} have shifted towards data-driven methods. These methods leverage large-scale driving datasets \cite{caesar2020nuscenes, wilson2023argoverse, montali2023waymo} and behavior cloning to model traffic behaviors. Despite these advancements, existing methods still focus on predicting marginal open-loop trajectories for individual agents, potentially resulting in a lack of scene consistency. This means that when multiple agents' most likely trajectories are rolled out together, unnatural interactions and even collisions can occur \cite{chen2022scept}, compromising the realism and reliability of simulated scenarios. 


Furthermore, traffic simulators should offer versatility and enable controllable scenario generation to satisfy both explicit (e.g., goals or targets) and implicit (e.g., collision avoidance) requirements. Existing models are often specialized in generating nominal behaviors conditioned on high-level objectives, such as navigation goals \cite{zhang2023trafficbots}. However, scenario generation models should also address the long-tail challenge in most datasets, where rare and dangerous situations are underrepresented \cite{liu2024curse}, potentially leading to inadequate testing in critical scenarios. 
Previous works in safety-critical traffic scenario generation \cite{cao2022advdo, hanselmann2022king, rempe2022generating, zhang2023cat} have employed tailored training objectives to encourage adversarial behaviors (e.g., inducing collisions with the ego agent), which limits their ability to capture the nuances of real-world traffic interactions. For instance, a model trained to generate aggressive driving behaviors may struggle to produce scenarios where surrounding agents attempt to avoid such adversarial actions, highlighting the constraints imposed by its specialized learning objectives. 
Therefore, a more generalized traffic simulation framework is needed to produce a wide spectrum of scenarios accommodating various requirements at inference time without retraining. This flexibility is crucial in traffic simulation, as it allows users to assign different behavioral modes to agents, enables the evaluation of edge cases, and facilitates the exploration of long-tail situations, providing a more comprehensive and customizable simulation tool. 



To overcome these limitations, we leverage diffusion models \cite{sohl2015deep, song2019generative, song2020score, ho2020denoising}, a class of generative modeling methods that gradually recover structured data from random noise. We develop \textbf{\model} (\modelabb), a flexible and scalable model for traffic behavior generation. \modelabb{} is capable of generating realistic, interactive, and controllable traffic scenarios for a large number of agents. As illustrated in \cref{fig:0}, despite being trained only on commonly used public road traffic datasets, \modelabb{} can accommodate a variety of tasks, such as normal or adversarial behavior generation, counterfactual reasoning, and reactive simulation, through guided sampling combined with user-specified objectives. Our VBD model demonstrates strong performance on the Waymo Sim Agent benchmark and offers versatile generation capabilities. Our main contributions are summarized as follows:
\begin{enumerate}
\item We propose \textbf{\model} for generalized traffic simulation, capable of generating realistic and controllable traffic agent behaviors with strong closed-loop simulation performance.
\item We demonstrate the versatility of \modelabb{} in simulations, showcasing its ability to generate diverse user-specified scenarios through flexible guidance schemes compatible with optimization objectives, behavior priors, and game-theoretic structures.
\item We conduct ablation studies to investigate the effects of various training and inference settings on multi-agent behavior generation.
\end{enumerate}


%% file: sections/related_works.tex
\section{Related Work}
\subsection{Traffic Simulation}
Traditional model-based or heuristic-based methods \cite{treiber2000congested, kesting2007general} often fail to capture complex agent interactions in real-world scenarios, resulting in a significant gap between simulation and reality. Consequently, there has been a growing shift towards learning-based methods to enhance the realism and interactivity of traffic simulations \cite{suo2021trafficsim}. 
For example, BITS \cite{xu2023bits} employs imitation learning to simulate agent behaviors by inferring high-level intentions and replicating low-level driving actions. The socially-controllable behavior generation model proposed in \cite{chang2023editing} focuses on simulating social and interactive behaviors. Symphony \cite{igl2022symphony} integrates learning-based policies with parallel beam search to further enhance realism. TrafficBots \cite{zhang2023trafficbots} introduces a shared policy conditioned on specific goals to generate configurable behaviors. Trajeglish \cite{philion2023trajeglish} introduces a multi-agent sequence of motion tokens using a GPT-like encoder-decoder architecture. Another research direction focuses on generating safety-critical or adversarial scenarios to test the robustness of driving systems. STRIVE \cite{rempe2022generating} generates challenging scenarios that can induce collisions with the ego planner through optimization in latent space. Similarly, AdvDO \cite{cao2022advdo} and KING \cite{hanselmann2022king} utilize optimization-based methods to generate adversarial trajectories for robust planning. CAT \cite{zhang2023cat} selects conflicting trajectories from agents' predicted behavior distributions to construct adversarial scenarios, supporting robust training of RL driving agents. However, most existing simulation models lack versatility, as they are specifically trained either for normal behaviors (maximum likelihood) or adversarial scenarios. We aim to bridge this gap by developing a generative framework that supports various tasks and requirements to enable generalizable, controllable, and realistic traffic simulation.

\subsection{Behavior Prediction}
Behavior or trajectory prediction models are widely utilized in both traffic simulations and driving policies \cite{zhang2023trafficbots, sun2023drivescenegen, feng2023trafficgen}. Recent advances in learning-based behavior prediction models have significantly improved the accuracy of predicting single-agent motions \cite{huang2022multi, nayakanti2023wayformer, zhou2023query} as well as joint multi-agent interactions at the scene level \cite{mo2022multi, shi2023mtr++, Huang_2023_ICCV}. Leveraging large amounts of real-world data, these models are capable of generating multi-modal distributions of possible behaviors for multiple agents in a scene. 
However, these prediction models often face limitations in scalability and efficiency when applied to large-scale scenarios with numerous agents. Therefore, more powerful generative models, including diffusion models \cite{jiang2023motiondiffuser, niedoba2023diffusion} and autoregressive Transformer models \cite{seff2023motionlm, zhou2024behaviorgpt}, have been increasingly applied in behavior prediction and generation tasks, demonstrating superior performance in large-scale predictions and simulations. Our proposed VBD model integrates multi-modal behavior prediction as high-level intention priors, enabling the estimation of realistic behavior distributions and sophisticated control over target agents' behaviors. This integration facilitates the diffusion model in generating specific scenarios based on the intended behaviors of particular agents. 

\subsection{Diffusion Models for Traffic Simulation}
Diffusion models \cite{song2019generative,ho2020denoising, song2020score, song2020denoising} have gained widespread popularity in a variety of generative tasks, including image \cite{zhang2023adding}, audio \cite{kong2020diffwave}, and video \cite{esser2023structure} generation. Recently, their application has been extended to traffic scenario generation.
For instance, SceneDM \cite{guo2023scenedm} utilizes a diffusion model to generate consistent joint motions for all agents in a scene. MotionDiffuser \cite{jiang2023motiondiffuser} employs a diffusion-based representation for joint multi-agent motion prediction and introduces a constrained sampling framework for controlled trajectory sampling. CTG \cite{zhong2023guided} combines diffusion modeling with signal temporal logic (STL) rules to ensure compliance with traffic rules in generated trajectories. CTG++ \cite{zhong2023language} leverages Large Language Models (LLMs) to translate user queries into loss functions, guiding the diffusion model toward generating query-compliant scenarios. TRACE \cite{rempe2023trace} proposes a guided diffusion model to generate future trajectories for pedestrians, employing analytical loss functions to impose trajectory constraints. Other models, such as DiffScene \cite{xu2023diffscene} and Safe-Sim \cite{chang2023controllable}, adopt guided diffusion techniques with adversarial optimization to generate safety-critical traffic scenarios. SceneDiffuser \cite{jiang2024scenediffuser} proposed to use amortized diffusion to improve the quality and computational efficiency for closed-loop rollouts. 
While our work is closely related to these guided diffusion models, these approaches often remain limited to specific tasks and lack versatility across different simulation applications.

%% file: sections/method/problem_formulation.tex
\section{Preliminaries} 
\label{sec: prob_form}

\subsection{Problem Formulation} 
We formulate traffic scenario generation as a conditional trajectory optimization problem in a multi-agent setting. A scenario $\scenario = (\traj, \ctrlseq, \condition)$ consists of agent trajectories $\traj = (\traj^1,\dots,\traj^\nagents) \in \reals^{\nagents \times \thorizon \times \nx}$, corresponding control sequences $\ctrlseq = (\ctrlseq^1,\dots,\ctrlseq^\nagents) \in \reals^{\nagents \times \thorizon \times \nc}$, and scene context $\condition \in \reals^{D_c}$ that encodes maps, traffic signals, and initial agent states $\state_0$.

We can define the scenario generation task as a finite-horizon optimal control problem:
\begin{equation}
\begin{split}
\underset{\ctrlseq}{\min} ~& \cost_\hparam(\traj, \ctrlseq; \condition),\\
\text{s.t.}\quad & \traj_0 = \state_0, \\
& \traj_{\tdisc+1} = \dyn(\traj_\tdisc, \ctrlseq_\tdisc),\ \forall \tdisc \in {0,\dots,\thorizon-1}
\end{split}
\label{eqn:opt}
\end{equation}
where $\dyn$ defines discrete-time joint dynamics, and $\cost_\hparam$ promotes realistic, feasible behaviors.

To model the distribution over control sequences $\ctrlseq$ conditioned on context $\condition$, we adopt the Maximum Entropy IRL \cite{ziebart2008maximum} formulation, representing the target distribution as a Boltzmann distribution:
\begin{equation}
\distrdata(\ctrlseq|\condition) \approx \distr_\theta(\ctrlseq|\condition) := \frac{1}{Z_\theta} \exp(-\cost_\hparam(\traj(\ctrlseq), \ctrlseq;\condition)),
\label{eqn:ebm_merge}
\end{equation}
which resembles an Energy-Based Model (EBM) \cite{lecun2006tutorial, song2021train}. The learning objective is to maximize the conditional likelihood of $\ctrlseq$ under the model $\distr_\theta$:
\begin{equation}
\theta = \argmax_{\hat\theta}~\expectation_{\scenario \sim \dataset}[\log \distr_{\hat\theta}(\ctrlseq|\condition)].
\label{eqn:cmle_merge}
\end{equation}

Directly estimating $\nabla_{\ctrlseq} \log \distr_\theta(\ctrlseq|\condition)$ (the score function) is challenging due to limited coverage in low-likelihood regions. To address this, recent works have employed diffusion models \cite{ho2020denoising, song2020score, song2020denoising}, which gradually transform data samples into noise through a forward stochastic process and then learn to reverse the process. Training such a model can be interpreted as learning the gradient steps of trajectory optimization in the score space, thus bridging generative modeling and control. As shown by \cite{du2019implicit, liu2022compositional, chi2023diffusion}, reverse diffusion sampling in this context approximates stochastic gradient descent toward low-cost, high-likelihood trajectories. This insight allows diffusion models to serve as both generative models and implicit planners that solve the control problem defined in \cref{eqn:opt}.

\subsection{Denoising Diffusion Probabilistic Model (DDPM)}
The discrete-time formulation of the forward diffusion process of DDPM \cite{nichol2021improved} can be described as:
\begin{equation}
q(\tilde{\ctrlseq}_1, \cdots, \tilde{\ctrlseq}_K | {\ctrlseq}) := \prod_{k=1}^K q(\tilde{\ctrlseq}_k | \tilde{\ctrlseq}_{k-1}),   
\end{equation}
\begin{equation}
q(\tilde{\ctrlseq}_{k} | \tilde{\ctrlseq}_{k-1}) := \mathcal{N} \left(\tilde{\ctrlseq}_{k}; \sqrt{1-\beta_k} \tilde{\ctrlseq}_{k-1}, \beta_k \mathbf{I} \right),  \label{eqn: diffusion_forward}
\end{equation}
where $\beta_k \in (0, 1)$ is the $k$-th noise scale from a predefined noise scheduling, ${\ctrlseq}$ is the clean control action sampled from data distribution, and $\tilde{\ctrlseq}_k$ is the noisy action samples at diffusion step $k$. In the final step $K$, the data distribution approaches an isotropic Gaussian distribution $q(\tilde{\ctrlseq}_K) \approx \mathcal{N}(\tilde{\ctrlseq}_K; 0, \mathbf{I})$. Let $\alpha_k = 1-\beta_k$ and $\bar\alpha_k = \prod_{i=0}^k\alpha_i$, we can directly sample $\tilde\ctrlseq_i$ from data $\ctrlseq$ without iterative diffusion via reparameterization trick:
\begin{equation}
\begin{split}
\tilde \ctrlseq_k &= \sqrt{\alpha_k}\tilde\ctrlseq_{k-1} +\sqrt{1-\alpha_k}\epsilon_{k-1} \\
&= \sqrt{\bar\alpha_k}\ctrlseq + \sqrt{1-\bar\alpha_k}\epsilon, \  \epsilon \sim \mathcal{N}(0, \mathbf{I}).
\end{split}
\label{eqn: ddpm_forward_sample}
\end{equation}

The generation process is accomplished by learning to reverse the forward diffusion process based on context information $\condition$. The reverse diffusion process starts with an isotropic Gaussian noise $q(\tilde{\ctrlseq}_K)$ and can be expressed as:
\begin{equation}
 p_\theta(\tilde{\ctrlseq}_0, \cdots,\tilde{\ctrlseq}_K | \condition) := q(\tilde{\ctrlseq}_K) \prod_{i=1}^K p_\theta(\tilde{\ctrlseq}_{k-1} | \tilde{\ctrlseq}_{k}, \condition), 
\end{equation}
\begin{equation}
p_\theta(\tilde{\ctrlseq}_{k-1} | \tilde{\ctrlseq}_{k}, \condition) := \mathcal{N} \left( \tilde{\ctrlseq}_{k-1} ; \mu_\theta (\tilde{\ctrlseq}_k, \denoiser(\tilde\ctrlseq_k, k, \condition)), \sigma^2_k \mathbf{I} \right),
\label{eqn: step}
\end{equation}
where $\mu_\theta$ calculates the posterior mean of the noise at $k-1$ step from $\tilde\ctrlseq_k$ and DDPM denoiser output $\denoiser(\cdot)$, and $\sigma_k$ is the standard deviation according to the fixed noise schedule. 

Specifically, the denoiser $\denoiser(\cdot)$ estimates the clean action trajectory sample $\hat{\ctrlseq}_k$ from the current noisy sample $\tilde{\ctrlseq}_k$, according to which the mean of the previous noisy sample $\tilde{\ctrlseq}_{k-1}$ can be derived as follows:
\begin{equation}
\label{mean}
 \mu_{k} :=  \frac{\sqrt{\bar \alpha_{k-1}} \beta_k}{1 - \bar \alpha_k} {\hat{\ctrlseq}_k} + \frac{\sqrt{\alpha_k} (1 - \bar \alpha_{k-1})}{1 - \bar \alpha_k} \tilde{\ctrlseq}_{k}.    
\end{equation}

\begin{figure*}[htp] 
    \centering
    \includegraphics[width=0.98\linewidth]{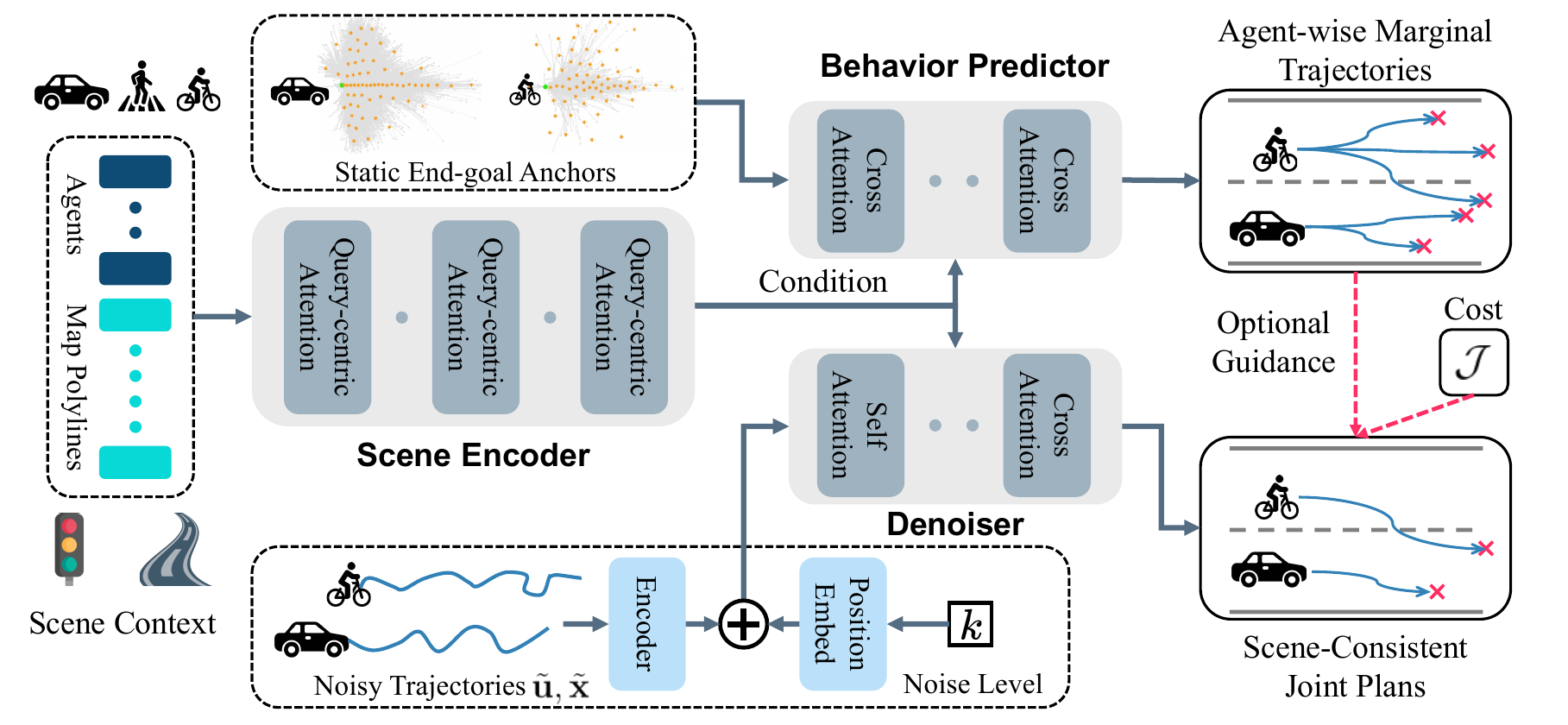}
    \caption{Overview of the proposed VBD model structure. The input scenario context is encoded through a query-centric Transformer scene context encoder. The behavior predictor generates marginal multi-modal trajectories. The denoiser predicts the joint multi-agent future trajectories while attending to themselves and the condition tokens. During inference, the predicted behavior priors or user-defined model-based objectives $\cost$ can be used to guide the denoising process to generate desired scenarios.}
    \label{fig: model_structure}
\end{figure*}

\subsection{Controllable and Compositional Generation} 
We also want to generate scenarios that satisfy a specific user requirement $y$ without retraining the model. For example, $y$ can be defined as the goal or the reference path for individual agents, or it can describe a soft constraint, such as obeying the speed limit or avoiding collisions. We modify the optimization objective to:
$\cost_\hparam(\traj,\ctrlseq;\condition)+\cost_y(\traj,\ctrlseq;\condition)$. Plugging into the EBM representation, we obtain a new conditional distribution: 
$\distrdata(\ctrlseq|\condition, y)\propto \distrdata(\ctrlseq|\condition)\distrdata(y|\ctrlseq,\condition)$, 
where $\distrdata(\ctrlseq|\condition)$ is the data distribution we approximated through generative modeling and $\distrdata(y|\ctrlseq,\condition)$ is the likelihood of $y$. This immediately resembles the compositionality in EBM \cite{du2019implicit} and Classifier Guidance in diffusion models \cite{dhariwal2021classifierguided}. Specifically, we can sample the reverse process with a conditional score function:
\begin{equation}
\begin{split}
    \nabla_{\tilde\ctrlseq} \log\distrdata_k(\tilde\ctrlseq|\condition, y) & \approx\score_\theta(\tilde{\ctrlseq}|\condition,y,k)  \\ 
    & = 
    \score_\theta(\tilde{\ctrlseq}|\condition,k) + \nabla_{\tilde\ctrlseq} \log\distrdata_k(y|\tilde\ctrlseq, \condition), 
\end{split}
\end{equation}
where $\distr_k(y|\tilde\ctrlseq, \condition)$ is the likelihood of $y$ given the noised action $\tilde\ctrlseq$ at step $k$. It is important to note that $\distr_k(y|\tilde\ctrlseq, \condition)$ is not equivalent to the likelihood of $y$ in the data distribution $\distrdata(y|\ctrlseq, \condition)$, therefore it is typically required to train a separate model \cite{dhariwal2021classifierguided, janner2022planning}. However, we can utilize a practical approximation to the gradient of the noised likelihood with the gradient of $\cost_y$ for guidance, which enables flexible composition and controllability with additional objectives without training. 

In the reverse diffusion process, the guidance is implemented by subtly altering the predicted mean of the model at each denoising step \cite{zhong2023guided, zhong2023language}. Initially, we can directly approximate $\log\distrdata_k(y|\tilde\ctrlseq, \condition)\approx \log\distrdata(y|\tilde\ctrlseq, \condition) = \cost_y(\traj(\tilde\ctrlseq),\tilde\ctrlseq;\condition)$; therefore, the denoising mean is modified to impose guidance in the form of a score function $\mathcal{J}_y$ as:
\begin{equation}
\tilde \mu_k = \mu_k + \lambda \sigma_k \nabla_{\mu_k} \mathcal{J}_y(\mu_k),
\end{equation}
where parameter $\lambda$ controls the strength of the guidance.

However, calculating the gradient based on the mean of noisy actions is difficult and inaccurate, and thus manipulating the noise mean using the noisy gradient can result in errors and instability. To address this, we utilize an alternative approach \cite{jiang2023motiondiffuser} to approximate $\log\distrdata_k(y|\tilde\ctrlseq, \condition)\approx \cost_y(\traj(\denoiser(\tilde\ctrlseq)),\denoiser(\tilde\ctrlseq);\condition)$. This means calculating the objective function using the one-step generation result from the denoiser rather than the noisy mean. The modification in the denoising step is expressed as:
\begin{equation}
\tilde \mu_k = \mu_k + \lambda\sigma_k \nabla_{\tilde{\ctrlseq}_k} \mathcal{J}_y(\mathcal{D}_\theta (\tilde{\ctrlseq}_k)),
\end{equation}
where gradient $\nabla_{\tilde{\ctrlseq}_k}$ is calculated with respect to the noisy actions $\tilde{\ctrlseq}_k$ and necessitates differentiation through the denoiser.

%% file: sections/method/model.tex
\section{Method}
\label{sec:model}

\subsection{Model Structure}
The \textbf{{\model}} (\modelabb) model consists of three main components, as illustrated in \cref{fig: model_structure}. The scene encoder $\mathcal{E}_\phi:\condition\mapsto\hat{\condition}$ encodes the scene context $\condition$ into its latent representation $\hat\condition$ using query-centric attention Transformers \cite{shi2023mtr++}. Leveraging rich scene context information from the encoder, the denoiser $\denoiser:(\hat\condition,\tilde{\ctrlseq},k)\mapsto \hat\ctrlseq$ directly predicts a joint clean control sequence $\hat\ctrlseq$ from $\hat\condition$ and noised control $\tilde\ctrlseq$ at step $k$. The behavior predictor $\predictor:(\hat\condition, \{\zeta^i\}_{i=1}^M) \mapsto \{\categorical_M(\hat\ctrlseq^{a}, \hat\omega^{a})\}_{a=1}^A$ predicts an $M$-mode \emph{marginal} categorical trajectory distribution of \emph{each agent} from $\hat\condition$ with the help of a set of representative static end-point anchors $\{\zeta^i\}_{i=1}^M$ extracted from data \cite{shi2023mtr++}. All three modules utilize a stack of query-centric self-attention and cross-attention blocks. 

\textbf{Scene Context}.
The scene conditions are divided into three categories: agents $\mathbf{c}_a \in \mathbb{R}^{A \times T_h \times D_a}$, map polylines $\mathbf{c}_{m, pl} \in \mathbb{R}^{M_l \times M_p \times D_p}$, and traffic lights $\mathbf{c}_{m, tl} \in \mathbb{R}^{M_t \times D_t}$. Here, $T_h$ denotes the number of historical steps, $M_l$ the number of polylines, $M_p$ the number of waypoints per polyline, $M_t$ the number of traffic lights, and $M_l+M_t=M$ represents the combined count of map elements. The feature sizes for agents, polylines, and traffic lights are represented by $D_a$, $D_p$, and $D_t$, respectively. 
The agent history tensor records each agent's historical state, including $x, y$ coordinates, heading angle ($\psi$), velocities ($v_x, v_y$), and bounding box dimensions ($l, w, h$), along with the agent type. Each map polyline, comprising $M_p$ waypoints, includes attributes such as $x, y$ coordinates, direction angle, the traffic light state controlling the lane, and lane type. The traffic lights tensor encompasses $x, y$ coordinates of stop points and the state of each traffic light. Before encoding, positional attributes of all elements are converted into their local coordinate systems; for agents, the reference point is their last recorded state, and for map polylines, it is the location of the first waypoint. 

\textbf{Scene Encoder}. 
We encode the agent history tensor utilizing a shared GRU network, which is then combined with the agent type embedding. For map polylines, an MLP is employed for encoding, followed by max-pooling along the waypoint axis; for traffic lights, we encode only their light status using an MLP. These tensors are then concatenated to form the initial scene encoding. The initial scene encoding is further processed using $L_{\mathcal{E}}$ query-centric Transformer layers to symmetrically encode the interrelationships among scene components, resulting in a uniform dimension for scene features $\hat{\condition} \in \mathbb{R}^{(A+M) \times D}$. In this approach, each scene element is translated into its local coordinate system and encoded with query-centric features, and the relative position of each pair of scene elements is calculated and encoded as edge attributes.

\textbf{Denoiser}.
The denoiser $ \mathcal{D}_\theta ({\tilde{\ctrlseq}}(k), k, \hat{\condition})$ receives as input the noised agent actions $\tilde{\ctrlseq}(k) \in \mathbb{R}^{A \times T \times 2}$ from the clean actions ${\ctrlseq}(0) = f^{-1}(\mathbf{x}(0))$, which are computed using an inverse dynamics model $f^{-1}$ from the ground truth states $\mathbf{x}(0) \in \mathbb{R}^{A \times T \times 4}$. Each agent's action at every timestep $u = [\dot v, \dot \psi]^T$ consists of acceleration and yaw rate, while the state comprises coordinates, heading, and velocity ($x, y, \psi, v$). The denoiser also takes the noise level $k$ and the encoded scene conditions $\hat{\condition}$ as inputs. It encompasses several Transformer layers with both self-attention and cross-attention mechanisms. 

Initially, the noised actions $\tilde{\ctrlseq}_k$ are converted to noised states $ \mathbf{\tilde x}(k) = f(\tilde{\ctrlseq}(k))$, encoded via an MLP, and concatenated with the noise level embedding. Then, a self-attention Transformer module is employed to model the joint distribution of future plans across agents. To maintain closed-loop rollout causality, a causal relationship mask \cite{huang2023dtpp} is used in the self-attention module, ensuring that information from future timesteps cannot be utilized at the current timestep. Furthermore, a cross-attention Transformer module is used to model the scene-conditional distribution by relating the noisy trajectories to the encoded scene conditions. This decoding block can be repeated several times, and the final embedding is fed into an MLP decoder to predict the denoised actions ${\hat \ctrlseq}(0)$ and clean states $\mathbf{\hat x}(0)$ through a dynamics function $f$. 

\textbf{Behavior Predictor}.
The behavior predictor $\predictor$ generates the marginal distributions of possible behaviors for individual agents by directly decoding from the encoded scene conditions $\hat{\condition}$. The predictor is composed of $L_{\mathcal{P}}$ cross-attention Transformer decoder layers. To accurately predict the probabilities of possible goals, we utilize type-wise static anchors (end goals) $\{\zeta^i\}_{i=1}^M$ extracted from data \cite{shi2023mtr++} as the modality query inputs to the initial Transformer decoder layer. The predictor iteratively refines its predictions through several decoding layers and finally generates $M$ trajectories over the future time horizon for each agent along with their associated scores, represented as $\{\categorical_M(\hat\ctrlseq^{a}, \hat\omega^{a})\}_{a=1}^A \in \mathbb{R}^{A \times M \times T \times 5}$. To ensure the kinematic feasibility of these goals, the model predicts action trajectories, which are converted into state trajectories using the same dynamics function $f$. Each waypoint in the state trajectory contains the state ($x, y, \psi, v$) and the probability of the trajectory $\hat\omega$.

\subsection{Model Training}
We implement a multi-task learning framework that concurrently trains the encoder, denoiser, and predictor components of our model. To train the denoiser, we aim to minimize the denoising loss:
\begin{equation}
    \label{eqn: denosie_loss_func}
\begin{split}
    \mathcal{L}_{\denoiser} = &\expectation_{\scenario\sim\distrdata,k\sim\uniform(0,K)}\expectation_{\tilde{\ctrlseq}\sim\distrdata_k(\cdot|\ctrlseq)} 
    \big[ \\
    &\lambda(k)\mathcal{SL}_1(
    \hat{\traj}(\denoiser(\hat\condition, \tilde\ctrlseq, k))-\traj) \big], 
\end{split}
\end{equation}
which is defined as the Smooth L1 loss between ground-truth trajectories $\traj$ and the trajectories $\hat\traj$ rolled out from $\hat\ctrlseq$. Essentially, the denoiser is trained to recover the clean trajectories under various noise levels. At each training step, noise level $k$ is sampled and applied to corrupt the ground-truth trajectories, and the denoiser is optimized to predict the denoised trajectories from the corrupted trajectories. Since the model predicts scene-level joint trajectories, all agent trajectories are affected by the same noise level. The design of the cost function plays a crucial role in generation performance.

In addition, we utilize a log noise schedule defined as:
\begin{equation} 
\bar \alpha_k = \frac{f(k)}{f(0)}, f(k) = \log \left( \frac{K+K\delta}{k+K\delta} \right), k=[0,1,\dots,K], \label{eqn: log_schedule}
\end{equation} 
where $K$ is the maximum diffusion step and $\delta$ is a scale factor that controls the variance changing rate of the log noise schedule. 
This schedule ensures that the signal-to-noise ratio (SNR) remains sufficiently low, preventing degradation in inference performance due to shortcut learning. Empirically, we found the performance of this noise schedule is significantly better than the Cosine noise scheduler \cite{nichol2021improved}. 



Training a denoiser with the scene encoder directly can be unstable, partially because the denoiser focuses on structured data from the noisy trajectories rather than the information from context encoding. To address this issue, we suggest incorporating an additional task in the model to predict marginal multi-modal trajectories, which can more effectively attend to the context encoding. This setting not only stabilizes training and enhances the overall learning performance but also provides behavior priors for individual agents. To train the behavior predictor $\predictor$, we follow the multi-trajectory-prediction (MTP) loss setting. This involves selecting the best-predicted mode $m^*$ that most closely matches the ground truth trajectory of each agent. To determine the best-predicted mode for an agent, the following criterion is applied:
\begin{equation}
m^* = \begin{cases} \arg \min_i ||\mathbf{ac}^i - \mathbf{x}_T||, & \text{if } \mathbf{x}_T \text{ is valid}, \\ \arg \min_i || \sum_t ({\mathbf{\hat x}}_t^i - \mathbf{x}_t) ||, & \text{otherwise}, \end{cases} 
\end{equation}
where $\mathbf{ac}^i$ is the static anchor point, $\mathbf{x}_t$ is the ground-truth point of the trajectory, and ${\hat x}_t^i$ is the predicted trajectory point. This means that if the ground-truth trajectory endpoint is invalid, the predicted trajectory with the smallest average displacement error is selected; otherwise, the trajectory corresponding to the closest anchor point is selected.

Subsequently, trajectories are chosen from the multi-modal predictions based on the indices $m^*$, and the Smooth $L1$ loss is computed between these selected trajectories and the ground-truth trajectories. For the training of the scoring function, cross-entropy loss is utilized, comparing the predicted logits with the given modes. The prediction loss is formulated as:
\begin{equation}
\begin{split}
\mathcal{L}_{\predictor} =  \expectation_{\scenario\sim\distrdata} \Bigg[ \sum_{a=1}^{A}\mathcal{SL}_1 \left(\hat\traj(\hat\ctrlseq^{a,m^*}) - \traj^a\right) \\+ \beta {CE} (m^*, \hat{\omega}^a) \Big],
\end{split}
\end{equation}
where $\beta$ is a hyperparameter. Note that this loss function is computed marginally, and time steps lacking ground-truth data or invalid agents are excluded from the loss calculation.

The total loss function for the multi-task learning model is formulated as:
\begin{equation}
\mathcal{L} = \mathcal{L}_{\denoiser} + \gamma \mathcal{L}_{\predictor},
\end{equation}
where $\gamma$ is a hyperparameter to balance different tasks.

\section{Experimental Setup}
\subsection{System Dynamics}

The diffusion model operates in the action space, and we assume that there is a dynamic function that can translate actions to physical states $\traj_{t+1} = f(\traj_t, \ctrlseq_t)$. We utilize unicycle dynamics as the system dynamics $\traj_{t+1} = f(\traj_t, \ctrlseq_t)$ to roll out actions into states. This model is shared across all types of agents, including vehicles, pedestrians, and cyclists. The current state of an agent is defined by its global coordinates $(x, y)$, yaw angle $\psi$, and velocities $v_x, v_y$. Given the action of an agent, including acceleration $\dot v$ and yaw rate $\dot \psi$, and the time length for one step $\Delta t$, the next-step state of the agent is calculated using the following forward dynamics $f$:
\begin{equation}
\begin{split}
x(t+1) & = x_t + {v_x}(t) \Delta t, \\
y(t+1) &= y_t + {v_y}(t) \Delta t, \\ 
\psi(t+1) &= \psi(t) + \dot \psi \Delta t, \\
v(t+1) &= \sqrt{{v_x}(t)^2 + {v_y}(t)^2} + \dot v \Delta t, \\
{v_x}(t+1) &= v(t+1) \cos \psi(t+1), \\
{v_y}(t+1) &= v(t+1) \sin \psi(t+1). 
\end{split}
\end{equation}

Since each operation in the dynamics function is differentiable, it can be integrated as a layer in the network to convert predicted actions into states.
Furthermore, we employ the inverse dynamics function $f^{-1}$ to calculate actions from ground-truth states, which is formulated as:
\begin{equation}
\begin{split}
\dot v(t) &= \frac{v(t+1) - v(t)}{\Delta t}, \\
v(t) &= \sqrt{v_x(t)^2 + v_y(t)^2}, \\
\dot \psi(t) &= \frac{\psi(t+1) - \psi(t)}{\Delta t}.
\end{split}
\end{equation}
Throughout our experiment, we normalize the acceleration by $1 \ \text{m/s}^2$ and the yaw rate by $0.15 \ \text{rad/s}$. 

{The unicycle dynamics used for mapping between actions and states may introduce small discrepancies between reconstructed and ground-truth trajectories. However, our empirical analysis shows that this effect is minor in practice. Across 40K agent trajectories from the Waymo Motion dataset, the average displacement error between 8-second reconstructed and ground-truth trajectories is 0.221 m, with a final displacement error of 0.511 m. These results show that the inverse dynamics model yields sufficiently accurate state-action reconstructions. In our framework, inverse dynamics is used only to obtain an approximate control-space representation, while supervision is applied on state trajectories after forward rollout through the differentiable forward dynamics function. This rollout-based supervision mitigates potential mismatches, ensuring consistent trajectory generation.}

\subsection{Dataset}
We conduct the experiments on the Waymo Open Motion Dataset (WOMD) \cite{Ettinger_2021_ICCV}, which comprises 486,995 nine-second logged real-world traffic scenarios for training and 44,097 scenarios for validation. The dataset provides trajectories for all agents and corresponding vectorized maps for each scenario. In the Waymo Sim Agents task \cite{montali2023waymo}, we evaluate our model's closed-loop simulation performance across 44,920 testing scenarios. In addition, we select 500 scenarios from the WOMD Validation Interactive split for additional experiments and ablation studies.

We choose only the WOMD instead of nuScenes \cite{caesar2020nuscenes} and Argoverse \cite{chang2019argoverse} for the following reasons. First, nuScenes and Argoverse are primarily designed for perception and short-term prediction benchmarks, which are insufficient for extended, interactive multi-agent simulation. Second, neither provides a standardized closed-loop evaluation framework that can simulate and replay the behavior of all traffic participants over the full duration of each scenario. In contrast, we use the Waymax environment \cite{gulino2023waymax} built on WOMD, which supports detailed, longer-horizon (8 seconds) closed-loop simulation of every agent in the scene, allowing us to rigorously assess multi-agent interaction dynamics.

\begin{figure*}[htp]
    \centering
    \includegraphics[width=\textwidth]{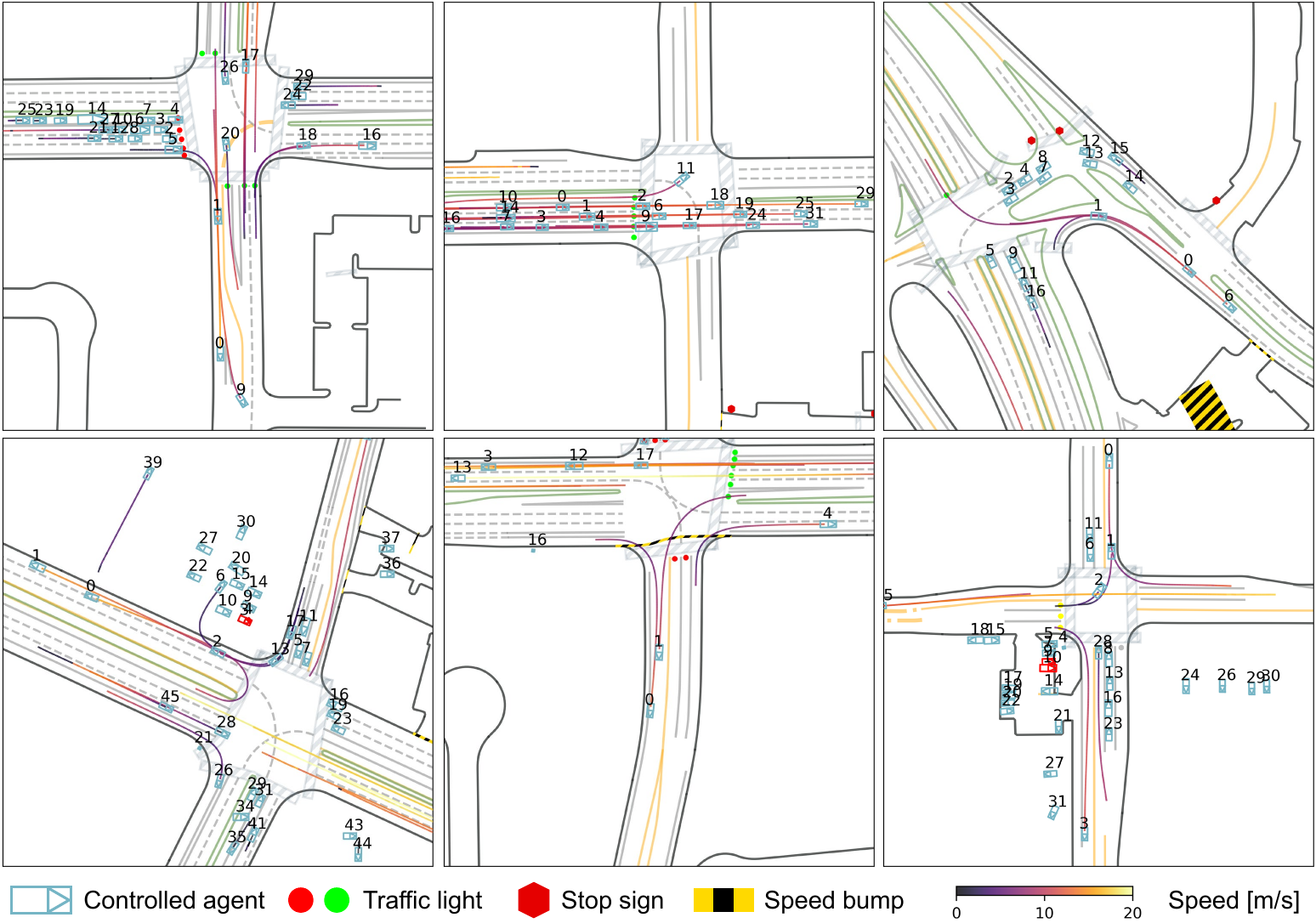}
    \caption{Performance of our VBD model on the Waymo Sim Agents task. The multi-agent diffusion policy is capable of controlling a large number of agents in an interactive and map-adherent manner for traffic simulation.}
    \label{fig: generation_ability}
\end{figure*}

\subsection{Implementation Details}

\textbf{Model Inputs}.
For the scene conditions, we consider $A=64$ agents, $M_l=256$ map polylines (each containing $M_p=30$ waypoints), and $M_t=16$ traffic lights in the scene. The flexibility of Transformer blocks allows {\modelabb} to be adapted to any number of agents during inference. {\modelabb} generates $\thorizon=80$ steps of future control sequences with step size $0.1 s$ based on only the current state and up to 1 s of past trajectories ($T_h = 11$). Empirically, we observe that discarding entire past trajectories and only keeping the current state during inference leads to better closed-loop performance by addressing the causal confusion issue in closed-loop testing \cite{cheng2024rethinking}. 

\textbf{Model Details}.
The scene encoder contains $L_{\mathcal{E}} = 6$ query-centric-attention Transformer layers, and the embedding dimension is $D=256$. The behavior predictor comprises $L_{\mathcal{P}}=4$ cross-attention Transformer layers and generates $64$ possible trajectories for each agent along with respective probability estimates. The denoiser includes two decoding blocks with four Transformer layers in total. Action sequences are effectively shortened to $T_f=40$ from $T$, by repeating actions over two time steps, which considerably reduces computational demands while ensuring high accuracy.

\textbf{Training Details}.
The proposed log variance schedule is adopted in the diffusion process, employing $K = 50$ diffusion steps, $\bar \alpha_{min}=1e-9$, and $\delta=0.0031$. 
For model training, the hyperparameters include a total loss function coefficient $\gamma=0.5$ and a predictor loss coefficient $\beta=0.05$. The model is trained using an AdamW optimizer with a weight decay of $0.01$. The initial learning rate is set at $0.0002$ and decays by $0.02$ every $1,000$ training steps, and a linear warm-up is employed for the first $1,000$ steps. The total number of epochs for training is $16$. Gradient clipping is implemented with a norm limit set to $1.0$. The training of the model utilizes BFloat16 Mixed precision on 8 NVIDIA L40 GPUs, with an effective batch size of $96$ across all GPUs.


%% file: sections/experiment.tex
\section{Closed-loop Traffic Simulation} 
\label{sec: exps_unguided}
\input{sections/exp_new/unguided_generation}

\section{Controllable Scenario Generation}
\label{sec: exps_guided}
\input{sections/exp_new/scenario_generation}

\section{Generation Beyond Data Distribution}
\label{sec: exps_safe}
\input{sections/exp_new/safety}

\section{Ablation Studies}
\label{sec: ablation}
\input{sections/exp_new/ablation}

%% file: sections/exp_new/unguided_generation.tex
We demonstrate the {\modelabb}'s capability and generality in generating interactive and realistic traffic scenarios through a closed-loop evaluation. For each scene, we initialize agent states from WOMD and leverage the Waymax simulator \cite{gulino2023waymax} for scenario roll-out. At each time of inference, {\modelabb} generates 8-second trajectories for all agents in the scene via the standard DDPM denoising scheme. All agents replan at 1 Hz using a receding horizon approach. 



We evaluate the scalability of {\modelabb}'s generation quality and interaction modeling capabilities using the Waymo Sim Agents Benchmark. In each scene, we are required to simulate 32 independent closed-loop rollouts of future scenarios for up to 128 agents over an 8-second horizon, conditioning on their trajectories from the preceding 1-second and the corresponding map contexts. Throughout the evaluation, {\modelabb} controls up to 64 agents surrounding the labeled self-driving vehicle in each scene, while a constant velocity policy controls other agents. By adopting this approach, we significantly reduce computational requirements when sampling millions of scenarios, leveraging the fact that most WOMD scenes contain fewer than 64 dynamic agents, with the remainder being stationary. We only utilize the current state of all agents by dropping out their past trajectories to overcome the causal confusion. We follow the official evaluation metrics of the Waymo Sim Agents benchmark \cite{montali2023waymo}, encompassing kinematic, interactive, and map-based metrics, and a meta realism metric is calculated as a weighted sum of these features.

\begin{table}[htp]
  \centering
  \caption{Testing Results on the 2024 Waymo Sim Agents Benchmark}
  \label{tab:wosac_results}
  \setlength{\tabcolsep}{1.5mm}
  \begin{tabular}{@{} 
      l| 
      c| 
      c| 
      c| 
      c 
    @{}}
    \toprule
    \textbf{Agent Policy} &
    \begin{tabular}[c]{@{}c@{}}
        \textbf{Realism Meta}\\
        \textbf{Metric} ($\uparrow$)
    \end{tabular} &
    \begin{tabular}[c]{@{}c@{}}
        \textbf{Kinematic}\\
        \textbf{Metric} ($\uparrow$)
    \end{tabular} &
    \begin{tabular}[c]{@{}c@{}}
        \textbf{Interactive}\\
        \textbf{Metric} ($\uparrow$)
    \end{tabular} &
    \begin{tabular}[c]{@{}c@{}}
        \textbf{Map-based}\\
        \textbf{Metric} ($\uparrow$)
    \end{tabular} \\
    \midrule
    SMART                   & 0.759 & 0.476  & 0.804 & 0.767  \\
    BehaviorGPT             & 0.747 & 0.433  & 0.800 & 0.764 \\
    GUMP                    & 0.743 & 0.478  & 0.789 & 0.740 \\
    MVTE                    & 0.730 & 0.450  & 0.771 & 0.743  \\
    SceneDiffuser           & 0.703 & 0.429  & 0.749 & 0.767  \\
    TrafficBotsV1.5         & 0.699 & 0.430  & 0.711 & 0.744  \\
    \textbf{VBD (Ours)}     & 0.720 & 0.417  & 0.782 & 0.721 \\
    \bottomrule
  \end{tabular}
\label{tab: wosac_results}
\end{table}

In \cref{tab: wosac_results}, we compare the performance of {\modelabb} with state-of-the-art generative models, including BehaviorGPT \cite{zhou2024behaviorgpt}, SMART \cite{wu2024smart}, and GUMP \cite{hu2024solving}, behavior cloning models, such as MVTE \cite{wang2023multiverse} and TrafficBotsV1.5 \cite{zhang2024trafficbots}, and concurrent diffusion-based model SceneDiffuser \cite{jiang2024scenediffuser}.
We demonstrate that the simulation performance of the {\modelabb} model is comparable to state-of-the-art generative models and outperforms other motion prediction models. In addition, we observe better simulation realism metrics compared to the SceneDiffuser \cite{jiang2024scenediffuser} baseline. Notably, our {\modelabb} model achieves this with fewer parameters than autoregressive generation models, achieving a balance between performance and computational efficiency. We present a selection of qualitative simulation results in \cref{fig: generation_ability}, showcasing the model's ability to generate interactive and realistic traffic scenarios. The results in the Waymo Sim Agents benchmark suggest that our VBD model is reliable at generating realistic agent interactions through its joint multi-agent diffusion policy. Closed-loop simulation videos can be found on the \href{https://sites.google.com/view/versatile-behavior-diffusion}{Project Website}.

{While the proposed VBD model achieves strong results on the Sim Agent benchmark, a performance gap remains compared to autoregressive models such as SMART and BehaviorGPT. We believe this discrepancy largely comes from architectural and training factors. In particular, the map representation and encoder design in VBD can be further improved to better capture spatial dependencies between the road map and agents’ future trajectories, and increasing the number of denoising steps yields limited benefit under the realism metric. Despite these factors, the VBD model offers distinct advantages, including joint multi-agent consistency and flexible inference-time controllability, enabling user-guided behavior generation such as adversarial maneuvers.}

%% file: sections/exp_new/scenario_generation.tex
A key strength of the {\modelabb} model lies in its capacity to generate customizable traffic scenarios without requiring retraining. This flexibility stems from the guided generation, allowing the model to adapt behaviors based on user-specified objectives, such as avoiding collisions or creating adversarial scenarios. By integrating these objectives into the denoising process, the {\modelabb} model provides a versatile and potent tool for simulating diverse traffic scenarios on demand. As outlined in \cref{alg:guide}, users can define cost objectives $\mathcal{J}_y$, enabling compositional optimization via the classifier guidance method \cite{dhariwal2021classifierguided}. This involves iteratively refining the noisy $\tilde\ctrlseq$ using both denoiser outputs and gradients of the form $\nabla{\tilde{\mu}_k} \mathcal{J}_y(\mathcal{D}\theta (\tilde{\mu}_k))$. Throughout this process, the denoiser ensures scenario realism, while the user's objective encourages the model to explore desired modes. Subsequent sections will demonstrate the application of {\modelabb}'s capabilities across various simulation tasks. Specifically, our guided generation tasks utilize $N_g=5$ gradient steps and a fixed scaling parameter $\lambda=0.1$ during all denoising operations.

\begin{algorithm}
\caption{Guided sampling with objective function}
\label{alg:guide}
\begin{algorithmic}[1]
\Require Denoiser $\mathcal{D}_\theta$, objective function $\mathcal{J}_y$, diffusion steps $K$, gradient steps $N_g$, scaling parameter $\lambda$, standard deviation $\sigma_k$
\State $\tilde{\ctrlseq}_K \sim \mathcal{N}(0, \mathbf{I})$ \Comment{Sample initial trajectory}
\For{$k \gets K$ to $1$}
    \State $\hat{\ctrlseq} \gets \mathcal{D}_\theta(\tilde{\ctrlseq}_k, k, \condition)$ \Comment{Predict denoised control sequence}
    \State $ {\tilde \mu}_k \gets \frac{\sqrt{\alpha_k}(1 - \bar{\alpha}_{k-1})}{1 - \bar{\alpha}_k} \tilde\ctrlseq_k + \frac{\sqrt{\bar{\alpha}_{k-1}}\beta_k}{1 - \bar{\alpha}_k} \hat{\ctrlseq}$    \Comment{Calculate unguided posterior $\tilde\mu_k$}
    \For {$i \gets 1$ to $N_g$}      
        \State $ {\tilde \mu}_k  \gets {\tilde \mu}_k + \lambda \sigma_k \nabla_{\tilde{\mu}_k} \mathcal{J}_y(\mathcal{D}_\theta (\tilde{\mu}_k))$ \Comment{Guidance gradient step}
    \EndFor
   
    \State $\tilde{\ctrlseq}_{k-1} \sim \mathcal{N}(\tilde \mu_k, \sigma_k^2 \mathbf{I})$ \Comment{Sample previous-step noised control sequence}
\EndFor
\State \textbf{Return:} Final control sequence ${\ctrlseq} \gets \tilde{\ctrlseq}_{0}$
\end{algorithmic}
\end{algorithm}

\begin{figure*}[t]
    \centering
    \includegraphics[width=0.99\linewidth]{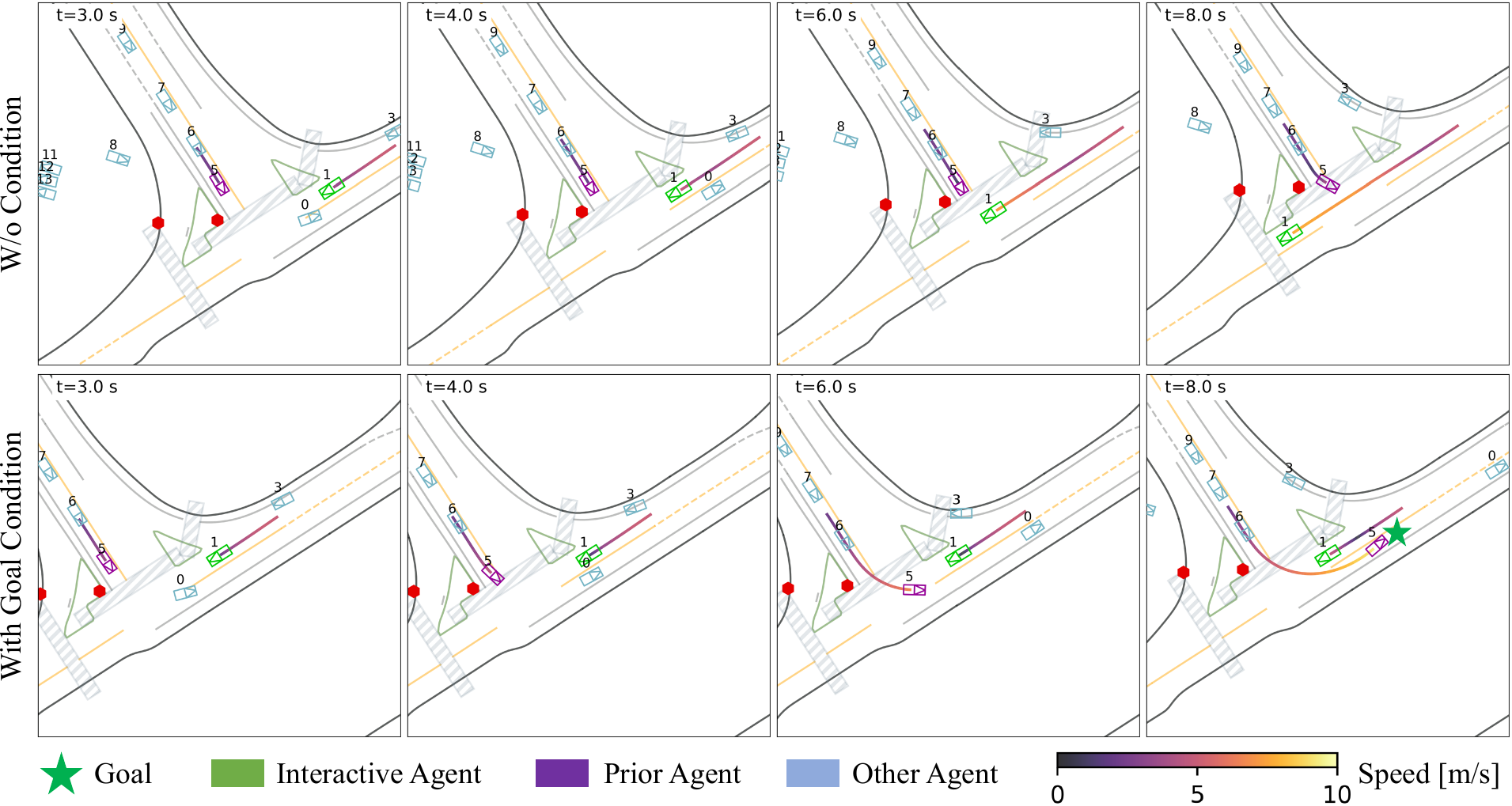}
    \caption{{\modelabb} produces scene-consistent traffic scenarios when target agents are conditioned on specific behavior priors. \textbf{Top}: Nominal {\modelabb} rollout without guidance generates a scene-consistent scenario, where Vehicle 5 (in purple) waits at the stop sign and then proceeds. \textbf{Bottom}: Using goal-guided diffusion to minimize Vehicle 5's final position w.r.t to a given goal, we enforce Vehicle 5 to run the stop sign. {\modelabb} model generates a scene-consistent scenario with Vehicle 1 yielding to Vehicle 5. }
    \label{fig: prior_1}
    \vspace{-0.4cm}
\end{figure*}

\begin{table*}[t]
  \caption{Simulation Results of Scenario Editing with Marginal Predictor and Diffusion Policy over Three Random Seeds}
  \label{tab: scene_consistency}
  \centering
  \begin{tabular}{@{}l|ccccc@{}}
    \toprule
    Method                                              & Collision [\%]  $\downarrow$                   & Off-road [\%] $\downarrow$                  & Wrong-way [\%] $\downarrow$                 & Kin. [\%] $\downarrow$                  & ADE [$m$] $\downarrow$\\ 
    \midrule
    Marginal Prediction              & 5.61$\pm$0.27            & 6.19$\pm$0.05          & 0.86$\pm$0.09          & 0.31$\pm$0.02           & 1.113$\pm$0.012 \\
    Diffusion                        & 2.47$\pm$0.09            & \textbf{1.21$\pm$0.14} & 0.57$\pm$0.02          & \textbf{0.24$\pm$0.01}  & 1.010$\pm$0.007 \\
    Post-Optimization via Diffusion  & \textbf{2.23$\pm$0.15}   & 1.26$\pm$0.10         & \textbf{0.52$\pm$0.11}  & 0.32$\pm$0.01           & \textbf{0.974$\pm$0.005} \\ 
  \bottomrule
  \end{tabular}
  \vspace{-0.4cm}
\end{table*}

We showcase the controllable scenario generation capabilities of our model in fulfilling various simulation tasks. To test this, we leverage the same subset of 500 WOMD interactive validation scenarios. Within each scene, we select the 32 agents closest to the labeled ego agent for evaluation, excluding all others. Our simulations span an $8$-second horizon with a replanning frequency of $1$ Hz, utilizing metrics from the Waymax simulator, including off-road incidents, collisions, wrong-way occurrences, kinematic infeasibilities, and log divergences. Furthermore, we analyze all quantitative results across three distinct random seeds, averaging outcomes to assess the generality and consistency.

We employ the following metrics to thoroughly test our proposed method quantitatively:
\begin{itemize}
\item \textit{Off-road}: A binary metric indicating whether a vehicle drives off the road, determined by its position relative to oriented road graph points. 
\item \textit{Collision}: A binary metric identifying if collisions happen between agents. 
\item \textit{Wrong-way}: A binary metric measuring whether a vehicle deviates from its intended driving direction. 
\item \textit{Kinematic infeasibility (Kin.)}: A binary metric assessing whether a vehicle's movement is kinematically plausible. 
\item \textit{Log divergence (ADE and FDE)}: These metrics quantify the deviation from ground truth behavior using displacement error. Average Displacement Error (ADE) is the L2 distance between an agent's simulated and ground-truth positions at each time step. Final Displacement Error (FDE) is the L2 distance error at $t = 8$ seconds. 
\item \textit{Minimum Log divergence (minADE and minFDE)}: These metrics quantify the minimum deviation from ground truth behavior using displacement error among all generated scenarios for each scene. 
\end{itemize}

\subsection{Composition with Behavior Priors}
Sampling diverse outputs from a conditional diffusion model is challenging, especially when denoising strongly relies on the context information \cite{sadat2023cads}. On the other hand, the behavior predictor in {\modelabb} captures the multi-modal trajectories of individual agents but will result in scene inconsistency if marginal trajectories are naively combined. This is because the predictor alone cannot ensure the coherence necessary for realistic multi-agent scenario generation. The denoiser in {\modelabb} can be used as an effective scenario optimizer and produce diverse and scene-consistent scenarios by first selecting goal positions for some target agents and generating joint multi-agent trajectories matching individual goals using guided sampling. Consider a scenario where we need to specify desired behaviors for $A^t$ agents, we can heuristically determine the target behaviors or goals for each target agent, represented by $\{\mathbf{g}^i\}_{i=1:A^t}$. The guidance function with behavior priors is:
\begin{equation}
\mathcal{J}_{goal} = -\sum_{i=1:A^t} \mathcal{SL}_1 (\mathbf{g}^{i} - \mathbf{x}_T^{i}),
\end{equation}
where $\mathcal{SL}_1$ denotes the Smooth L1 loss, $x_T^{i}$ is the state of an agent derived from actions using a differentiable dynamic function, and $T$ is the planning horizon. The other agents in the scene will not be directly influenced by the guidance.

\begin{figure*}[htp]
    \centering
    \includegraphics[width=0.98\linewidth]{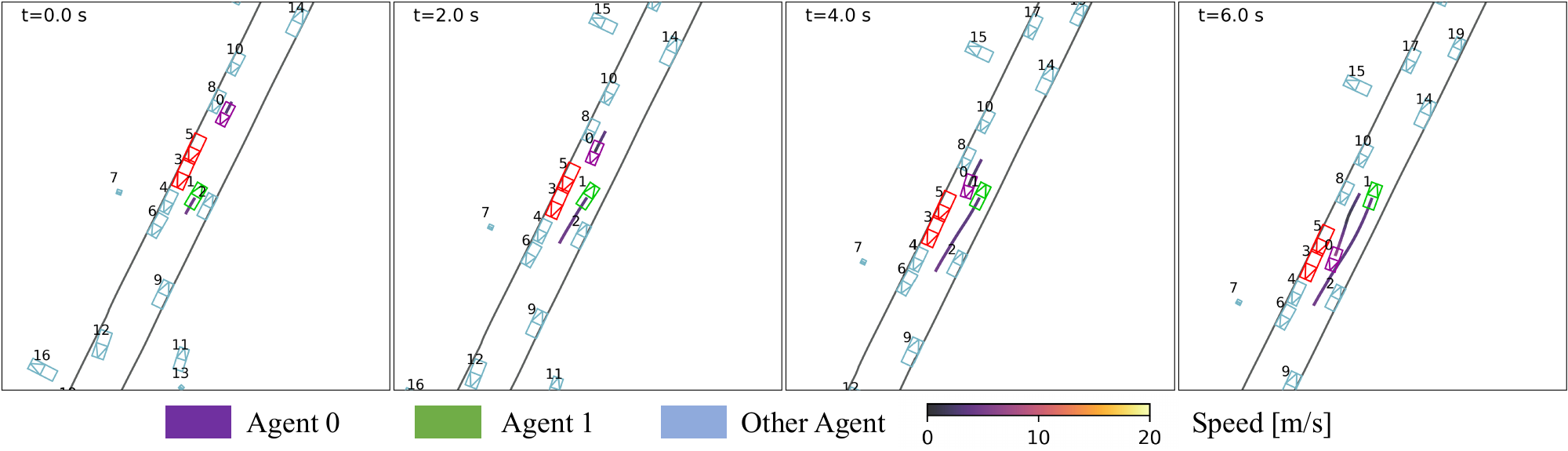}
    \caption{Composition of collision avoidance cost-based objective in diffusion can improve generation quality. Two vehicles dynamically interact and coordinate in a narrow passage scenario with collision cost function guidance. (Note: Vehicles 3 and 5 have been in collision since the initial step.)}
    \label{fig: interaction}
    \vspace{-0.3cm}
\end{figure*}

\begin{table*}[htp]
  \caption{Simulation Results of Collision Avoidance Cost-guided Diffusion over Three Random Seeds}
  \label{tab: condition}
  \centering
  \begin{tabular}{@{}l|ccccc@{}}
    \toprule
    Method                                              & Collision [\%]  $\downarrow$                   & Off-road [\%] $\downarrow$                  & Wrong-way [\%] $\downarrow$                 & Kin. [\%] $\downarrow$                 & ADE [$m$] $\downarrow$ \\
    \midrule    
    {\modelabb}                                           & 2.47$\pm$0.09            & 1.21$\pm$0.14          & 0.57$\pm$0.02          & \textbf{0.24$\pm$0.01}           & \textbf{1.010$\pm$0.007} \\ 
    {\modelabb} + Collision Avoid
    & \textbf{1.11$\pm$0.23}   &\textbf{1.15$\pm$0.17}  & \textbf{0.60$\pm$0.12} & 0.25$\pm$0.01                     & 1.113$\pm$0.010\\ 
  \bottomrule
  \end{tabular}
  \vspace{-0.3cm}
\end{table*}

\subsubsection*{Case Study 1: Scene-consistent Scenario Editing}
We evaluate the effectiveness of {\modelabb} as a scenario editing and optimization tool to generate scene-consistent interactions based on marginal behavior priors for target agents. Specifically, for each target agent, we select its most likely predicted trajectory as the goal. The baseline method (marginal prediction) directly executes the trajectory for target agents. For the post-optimization with diffusion, these trajectories serve as goal priors, and we employ guided sampling to minimize the mean $L2$ distance between the diffusion results and the selected goals.  
We compare their performance in generating specific behaviors (goal reaching) and test each method with three different random seeds. The results in \cref{tab: scene_consistency} indicate that the diffusion policy enhances scene consistency in the generated scenarios, given specific behaviors from marginal behavior priors. It significantly reduces the collision rate and better captures interactions between agents in the scenes for goal-reaching behaviors of target agents. Moreover, even when priors are selected from suboptimal samples, e.g., off-road or wrong-way, {\modelabb} can alleviate these cases and generate scenarios that conform to the scene context.

\cref{fig: prior_1} showcases the {\modelabb} model's capability in scene-consistent scenario editing, where one agent is assigned a target point based on marginal priors. In this setup, a human user oversees the simulation, manually selecting the target agent and its goal. By conditioning on the behavior that vehicle 5 makes a left turn, ignoring the stop sign, our {\modelabb} model's diffusion policy can generate a scenario such that vehicle 1 is compelled to yield to the other vehicle. Conversely, the nominal diffusion policy rollout without guidance follows a completely different but still scene-consistent traffic ordering.

\subsection{Composition with Cost Functions}
Incorporating collision avoidance guidance into the diffusion generation process can enhance the quality and realism of the generated scenarios and create nuanced, collision-avoiding behaviors among agents. This is implemented through a differentiable cost function, expressed as follows:
\begin{equation}
\mathcal{J}_{overlap} = \sum_{t=1}^T  \sum_{i,j}^A d_{ij}(\traj_t) \mathbbm{1}(d_{ij}(\traj_t) < \epsilon_d),
\end{equation}
where $d_{ij}$ represents the Minkowski distance between the footprints of agents $i$ and $j$ at time $t$. The parameter $\epsilon_d$ is the threshold for defining a potential collision. 

Further, to modify the driving behavior of a target agent (i.e., to make it more aggressive), we introduce a rush cost function that penalizes speed reduction, which is:
\begin{equation}
\mathcal{J}_{rush} = - \sum_{t=1}^T || a^{i}_t ||^2 \mathbbm{1}(a^{i}_t < 0),
\end{equation}
where $a^{i}_t$ represents acceleration action of agent $i$ at time $t$. By adding this function and the collision avoidance function together, we can implicitly modify the driving behavior of a specific agent within traffic flows.

\subsubsection*{Case Study 2: Enhancing Generation Quality} 
Since neural networks still exhibit unpredictability in generation, they may be prone to errors in some scenarios (e.g., off-road or collision behaviors). We aim to use a collision avoidance cost-based optimization objective in guided sampling to further improve the performance and generation quality of {\modelabb}. Specifically, we introduce a collision avoidance cost to maximize each agent's minimum distance from others across the horizon. 
As shown in \cref{tab: condition}, this collision avoidance objective effectively reduces the collision rate. 
The qualitative results in \cref{fig: interaction} further illustrate that incorporating a cost-based objective enables the model to generate collision-free interactions in the challenging narrow-passage scenario.

\subsubsection*{Case Study 3: Modifying Driving Style} 
\begin{figure*}[htp]
    \centering
    \includegraphics[width=0.99\linewidth]{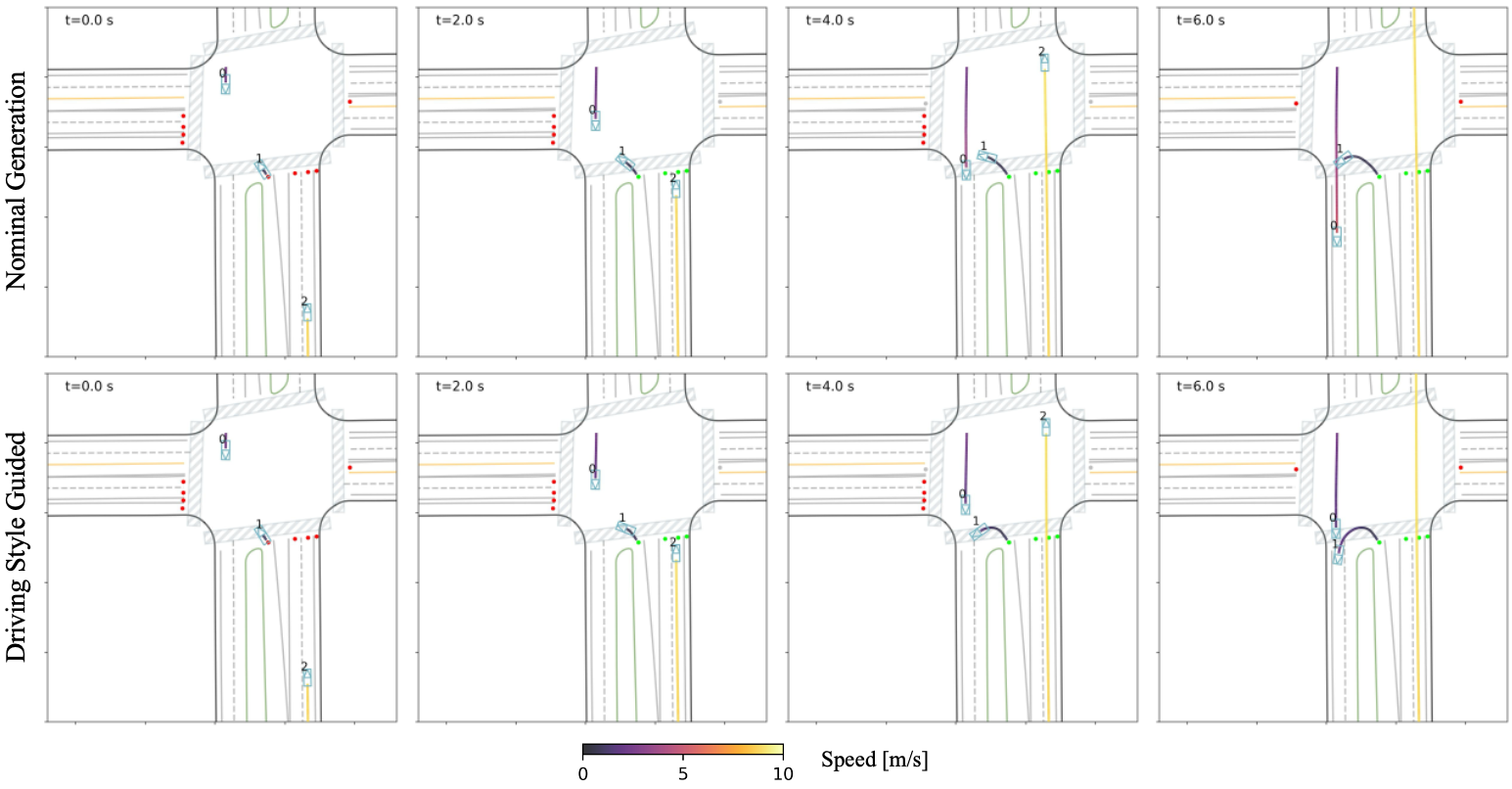}
    \caption{Composition of rush cost-based objective in diffusion can effectively change the driving style of an agent. \textbf{Top}: Nominal {\modelabb} rollout without guidance, where Vehicle 1 yields to Vehicle 2 before the U-turn maneuver. \textbf{Bottom}: Using rush cost-guided diffusion on Vehicle 1 makes it more aggressive, forcing Vehicle 2 to slow down and yield.}
    \label{fig: rush}
    \vspace{-0.3cm}
\end{figure*}

In simulation settings, it is often desirable to alter the driving behaviors of certain agents to generate different driving styles. To achieve this, we combine a collision avoidance cost and a rush cost to change the driving behavior of a target agent in the scene generation process using guided diffusion. This composite cost guidance can effectively change the behaviors of specific agents to show aggressive driving styles, while the {\modelabb} model ensures that surrounding agents remain responsive to these behavioral changes. We manually select a target agent in a scenario and consistently apply the guidance to change the driving style of the agent in closed-loop simulations. A qualitative example of modifying an agent's driving style is illustrated in \cref{fig: rush}, where we modify the driving style of Agent 1 to exhibit an aggressive U-turn maneuver. Concurrently, the {\modelabb} model generates scene-consistent behaviors for Agent 2, who adapts by yielding to the aggressive agent. 

%% file: sections/exp_new/safety.tex
A crucial use case of simulations is to create safety-critical scenarios to stress test the robustness of AV planning systems, which must be both realistic and reactive. We employ two guidance approaches (conflict prior guided diffusion and game-theoretic guided diffusion) in our {\modelabb} model to achieve this. Examples of the generated safety-critical and adversarial scenarios using our {\modelabb} model, implemented with these two composition methods, are showcased on the \href{https://sites.google.com/view/versatile-behavior-diffusion}{Project Website}. These examples demonstrate the efficacy of our methods in generating realistic scenarios that introduce collision risks for the AV, without requiring risk-specific driving datasets or any retraining or fine-tuning of the model. 

\begin{algorithm}[htp]
\caption{Game-theoretic-guided adversarial scenario generation with gradient descent-ascent}
\label{alg:game}
\begin{algorithmic}[1]
\Require Denoiser $\mathcal{D}_\theta$, diffusion steps $K$, evader index $e$, pursuer index $p$, guidance steps $N_g$, descent steps $N_e$, ascent steps $N_p$,
evader gradient mask $M_e$, pursuer gradient mask $M_p$,
scaling parameter $\lambda$, standard deviation $\sigma_k$.
\State $\tilde{\ctrlseq}_K \sim \mathcal{N}(0, \mathbf{I})$ \Comment{Sample initial trajectory}
\For{$k \gets K$ to $1$}
    \State $\hat{\ctrlseq} \gets \mathcal{D}_\theta(\tilde{\ctrlseq}_k, k, \condition)$ \Comment{Predict denoised control sequence}
    \State $ {\tilde \mu}_k \gets \frac{\sqrt{\alpha_k}(1 - \bar{\alpha}_{k-1})}{1 - \bar{\alpha}_k} \tilde\ctrlseq_k + \frac{\sqrt{\bar{\alpha}_{k-1}}\beta_k}{1 - \bar{\alpha}_k} \hat{\ctrlseq}$    \Comment{Calculate unguided posterior $\tilde\mu_k$}
    \For {$i \gets 1$ to $N_g$}
        \For {$i_d \gets 1$ to $N_e$}
        \State $\hat{\traj}\gets \dyn(\mathcal{D}_\theta (\tilde{\mu}_k))$
        \State $\cost\gets \min_t d_{e,p}(\hat{\traj}_t) + \min_t d_{r}(\hat{\traj}^e_t)$
        \State $ {\tilde \mu}_k^e  \gets {\tilde \mu}_k - \lambda \sigma_k M_e\nabla_{\tilde{\mu}_k^e} \cost $ \Comment{Gradient descent for evader}
        \EndFor
        \For {$i_d \gets 1$ to $N_p$}
        \State $\hat{\traj}\gets \dyn(\mathcal{D}_\theta (\tilde{\mu}_k))$
        \State $\cost\gets \min_t d_{e,p}(\hat{\traj}_t)+\min_t d_{r}(\hat{\traj}^p_t)$
        \State $ {\tilde \mu}_k^p  \gets {\tilde \mu}_k + \lambda \sigma_k M_p\nabla_{\tilde{\mu}_k^p} \cost$ \Comment{Gradient ascent for pursuer}
        \EndFor
    \EndFor
   
    \State $\tilde{\ctrlseq}_{k-1} \sim \mathcal{N}(\tilde \mu_k, \sigma_k^2 \mathbf{I})$ \Comment{Sample previous-step noised control sequence}
\EndFor
\State \textbf{Return:} Final control sequence ${\ctrlseq} \gets \tilde{\ctrlseq}_{0}$
\end{algorithmic}
\end{algorithm}

\subsection{Safety-Critical Scenario Generation via Conflict Priors}
We can utilize conflict-prior guidance to facilitate the generation of safety-critical scenarios. The behavior prior prediction significantly aids in identifying potentially unsafe agents, which is difficult for existing methods, and enhances the realism of the unsafe behaviors. By using the behavior predictor in our {\modelabb} model, we can obtain prior distributions of the possible movements of surrounding agents. From this distribution, agents that could conflict with the ego vehicle's plans are identified, and these selected priors are then used to guide the diffusion policy to generate expected scenarios. The process of selecting conflicting agents is represented as:
\begin{equation}
\mathbf{\hat x}^{adv} = \argmax_{i, j} \hat \omega_j^i \left[ col(\mathbf{\hat x}_j^i, \mathbf{\hat x}^{ego}_*) \mathbf{1}(\hat \omega_j^i > \epsilon_p) \right], \ \forall i \neq ego,   
\end{equation}
where $\mathbf{\hat x}_j^i, \hat \omega_j^i$ denote the trajectory and probability estimation for the prior mode $j$ of agent $i$, and $\mathbf{\hat x}_*^{ego}$ is the most-likely prior mode of the ego vehicle. $col$ is a collision probability function, which decreases linearly with the temporal proximity to a collision, and $\epsilon_p$ is a threshold controlling the possibility.

\begin{figure}[htp]
    \centering
     \includegraphics[width=\linewidth]{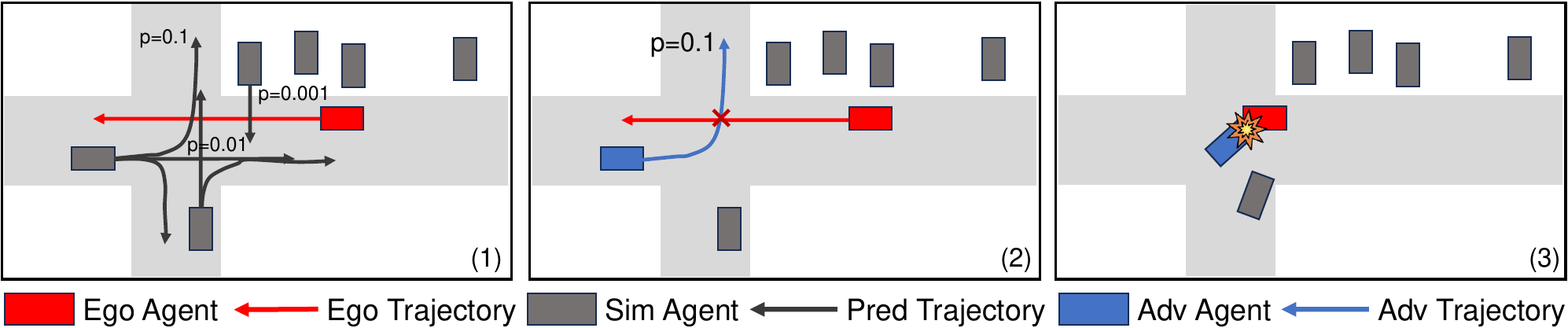}
     \caption{Illustration of conflict prior-guided safety-critical scenario generation. (1) The behavior predictor generates multi-modal trajectories for each sim agent; (2) From the prior distribution, select the agent behavior with the highest probability of conflicting with the most likely ego plan; (3) Generate the expected scenario using the diffusion policy.}
     \label{fig:safe}
     \vspace{-0.4cm}
\end{figure}

\begin{figure*}[htp]
    \centering
    \includegraphics[width=0.98\linewidth]{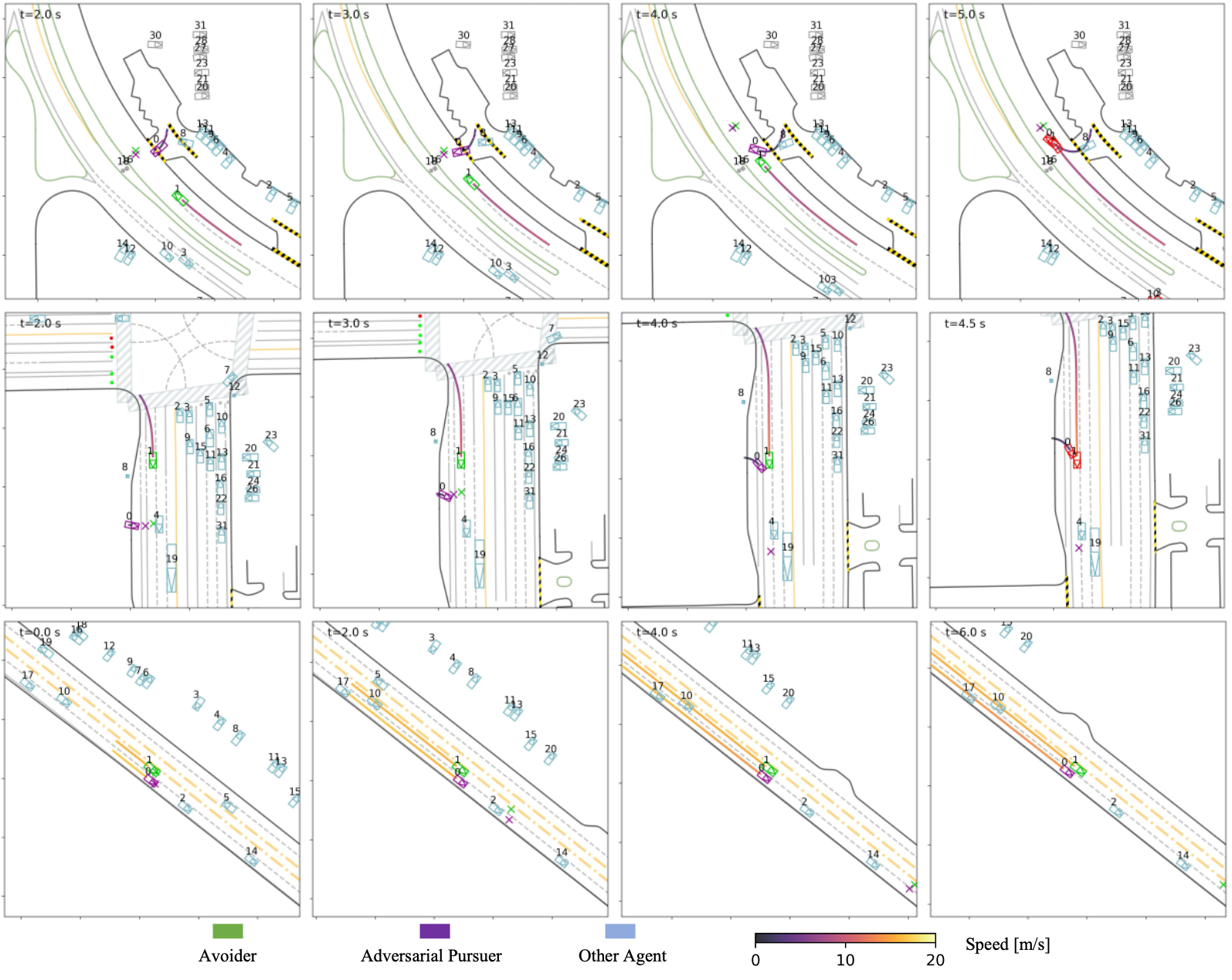}
    \caption{Safety-critical scenario generation through conflict priors guidance. \textbf{Top and Middle}: By selecting the highest posterior probability of adversarial agents' trajectories that conflict with the ego agent's most likely prediction, we can guide the denoising process to generate safety-critical scenarios. \textbf{Bottom}: Failure case of conflict prior guidance when there were no collision pairs from behavior priors.}
    \label{fig: conflict}
    \vspace{-0.4cm}
\end{figure*}

The core idea is to select the highest posterior probability of other agents' behaviors that conflict with the ego agent's normal driving behavior. The selected conflicting behaviors are then fed into the diffusion policy as guidance inputs. Meanwhile, the behaviors of other agents are controlled by the policy without guidance to respond to that situation. Notably, utilizing prior guidance in the diffusion policy results in more realistic safety-critical behavior compared to direct trajectory rollout, as well as robustness against selected priors that are less likely in real-world conditions. The proposed conflict prior guidance for safety-critical simulation is illustrated in \cref{fig:safe}, and generated safety-critical scenarios are presented in \cref{fig: conflict}.

\subsection{Safety-Critical Generation via Game-theoretic Guidance}
Conflict prior method is an effective sampling approach when agents' paths intersect with each other. However, this method may fail to capture the potential safety-critical scenario when no collision pairs can be automatically selected from behavior priors (e.g., the bottom row of \cref{fig: conflict}). 
To simulate interactions in more extreme conditions, such as adversarial driving behaviors, we model the interaction between two agents as a map-constrained pursuit-evasion game, where the pursuer aims to cause a collision with the evader and stay within the road boundary, while the evader attempts to avoid it. An effective approach to solving this game is through iterative best response (IBR), such as gradient descent-ascent (GDA). Specifically, we can apply $\tau$-GDA with time-scale separation to guarantee local convergence to a stable \textit{minimax equilibrium} of the pursuit-evasion game \cite{fiez2021global}. We update the pursuer more frequently than the evader to provide an information advantage to the adversarial agents.  


Generating a scenario with a game-theoretic solver often leads to unrealistic results. Instead, leveraging realistic traffic behaviors modeled by {\modelabb}, we propose a \textit{game-theoretic-guided diffusion} approach (\cref{alg:game}). This method involves alternating denoising steps, performing gradient descent and ascent for agents, and updating the noised $\tilde\ctrlseq$ until convergence. We can further enhance the realism of the scenario with \emph{optional} gradient masks $M_e$ and $M_p$, which allow us to adjust the adversity of the pursuer and the responsiveness of the evader by selectively updating gradients at certain timesteps.

\begin{figure*}[htp]
    \centering
    \includegraphics[width=\linewidth]{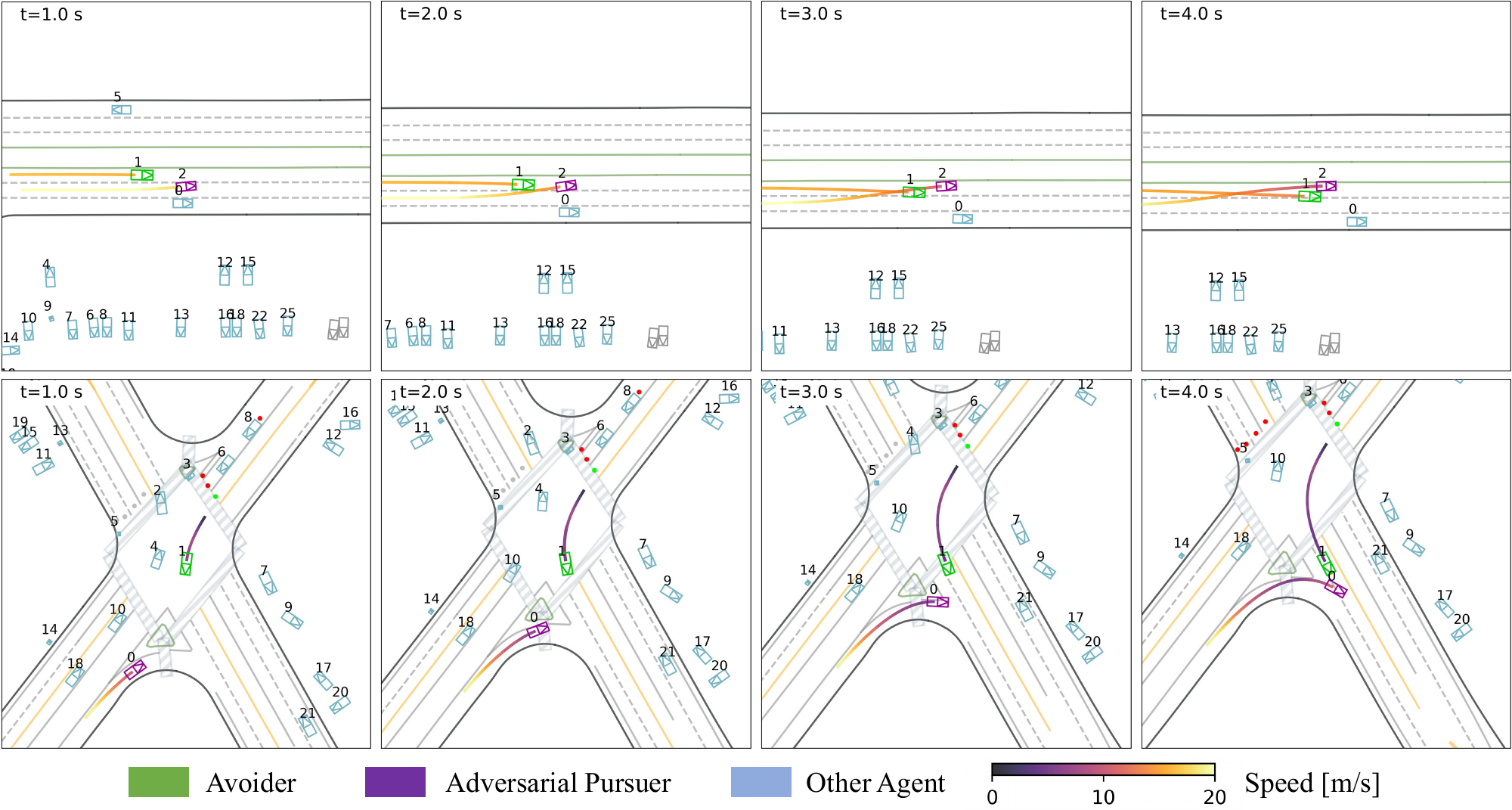}
    \caption{Results of game-theoretic adversarial behavior generation. \textbf{Top}: The adversarial pursuer merges in front of the evader, performs a brake check, and attempts to cause a rear-end crash. The evader immediately switches its lane and avoids the collision. \textbf{Bottom}: The adversarial pursuer merges aggressively into the adjacent lane, and the evader yields to the pursuer by slowing down.}
    \label{fig: safety_critic1}
\end{figure*}

We illustrate the generation of adversarial driving scenarios with the proposed game-theoretic guidance in various scenes in \cref{fig: safety_critic1}, where the proposed guidance can generate highly interactive maneuvers in both traffic merge and lane change scenarios.  
In contrast to previous works \cite{rempe2022generating, chang2023controllable, zhang2023cat}, which primarily sample adversarial behaviors based on fixed trajectories of the pursued vehicle, our approach proactively optimizes the actions of both the pursuer and evader. This results in highly interactive scenarios, enriching the realism of adversarial encounters. Our proposed method facilitates the generation of highly realistic scenarios, especially with regard to adversarial behavior, by ensuring that they are proportionately challenging and reactive to the maneuvers of the ego vehicle. This strategy overcomes the shortcomings of previous methods, which tend to generate unrealistically aggressive adversarial strategies.

%% file: sections/exp_new/ablation.tex

\begin{figure*}[htp]
    \centering
    \includegraphics[width=\linewidth]{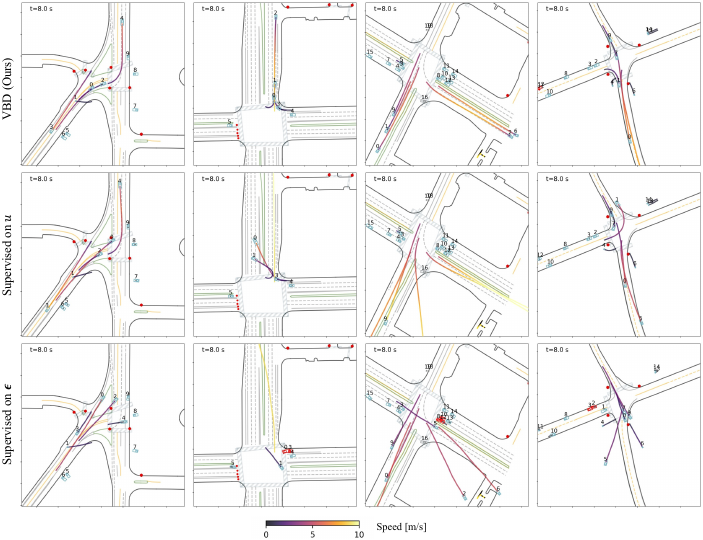}
    \caption{Comparison of training objectives in {\modelabb} by rolling out entire 8-second open-loop trajectories. The impact of training objectives is illustrated by examining the generated agent trajectories. \textbf{Bottom}: The model trained to predict noise $\epsilon$ fails to produce meaningful agent behaviors. \textbf{Middle}: The model trained to imitate control sequences $\ctrlseq$ fails to align with map features. \textbf{Top}: Supervised on roll-out trajectories, {\modelabb} significantly improves planning quality with the same architecture and training steps.}     
    \label{fig: loss_ablation}
\end{figure*}

\begin{figure*}[htp]
    \centering
    \includegraphics[width=0.95\textwidth]{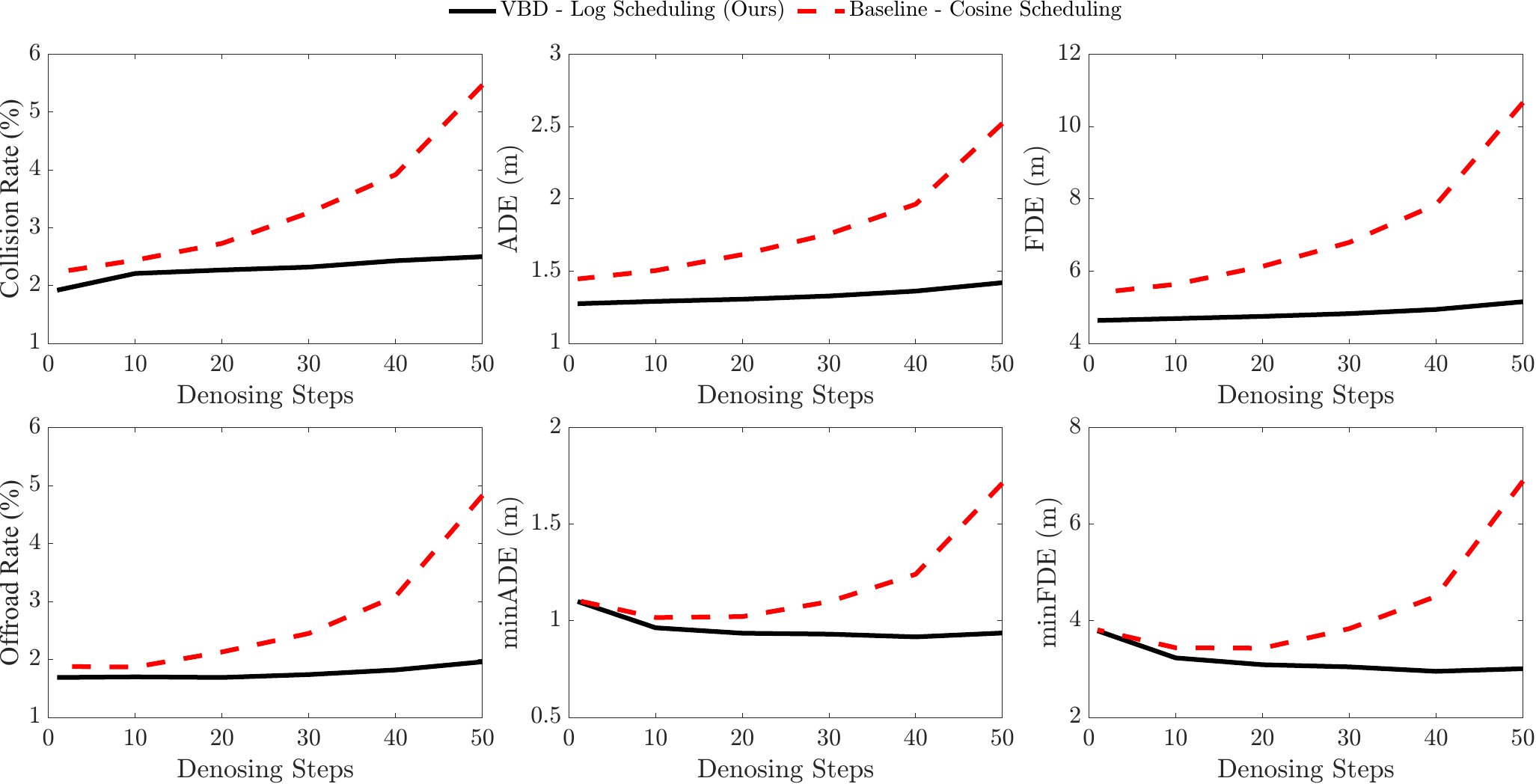}   
    \caption{Effects of noise scheduling on closed-loop simulation performance. The metrics (except for minADE and minFDE) are average values over 32 samples of closed-loop rollout. The {\modelabb} model trained with cosine scheduling exhibits significant performance degradation across all metrics. In contrast, {\modelabb} trained with the proposed log scheduling maintains low collision and off-road rates across all denoising steps. Additionally, minADE and minFDE metrics decrease as the number of denoising steps increases, reflecting the model's ability to generate scenarios closer to the ground truth. A slight increase in ADE and FDE metrics indicates improved sample diversity with more denoising steps.}
    \label{fig: schedule_results}
\end{figure*}

\subsection{Training Objective for Behavior Generation}
\label{sec: ablation_loss}
Traditional image generation tasks using diffusion models typically involve regressing either the injected noise or the ground truth image from a noised input. Notably, research has shown that predicting noise yields superior generation quality \cite{ho2020denoising}. However, our {\modelabb} model is trained with the loss function defined in \cref{eqn: denosie_loss_func}, which minimizes the $L_1$ distance between the trajectory rollouts derived from denoised action sequences and the corresponding ground-truth trajectories.

Our experimental results demonstrate that supervision through trajectory rollouts is crucial for generating high-quality traffic scenarios. Unlike image generation, where errors in pixel prediction do not compound, scenario generation involves sequential decision-making, where errors accumulate across time steps. Supervising either the control sequence $\ctrlseq$ or the noise $\epsilon$ in an open-loop manner fails to capture the temporal dynamics of traffic behaviors. To illustrate this, we trained a diffusion model with the same architecture as {\modelabb} for 20 epochs, supervising directly on the generated control sequences $\ctrlseq$ and the noise $\epsilon$. As shown in \cref{fig: loss_ablation}, models trained under these settings failed to produce coherent and realistic traffic scenarios, highlighting the necessity of trajectory rollouts as the training objective.

\subsection{Influence of History Encoding}
\label{sec: ablation_history}
Causal confusion presents a considerable obstacle when applying imitation learning-based models to closed-loop simulations. Consistent with prior studies \cite{de2019causal, cheng2024rethinking, bansal2018chauffeurnet}, our findings suggest that discarding historical information from agent representations and focusing solely on their current states substantially alleviates this problem. Specifically, we employ an aggressive dropout rate (50\%) for the past agent trajectories at training time and only supply the current joint state to the model during inference. 
As shown in \cref{tab: prediction}, experimental results reveal that models trained and tested with history dropout surpass those using full historical states encoding across all metrics, underscoring its effectiveness in mitigating causal confusion. 
Specifically, we observe that the baseline with full past trajectories overly accelerates or decelerates in closed-loop simulation and leads to significantly high displacement errors, highlighting its copying behavior through kinematic history.
By adopting history dropout, we achieve enhanced performance in closed-loop simulations, highlighting the significance of suitable state representation.

\begin{table*}[htp]
  \caption{Impact of History Encoding and Integrating Behavior Prior Prediction in VBD}
  \label{tab: prediction}
  \centering
  \begin{tabular}{@{}c|c|cccccc@{}}
    \toprule
    Predictor & No History  & Collision [\%] $\downarrow$   & Off-road [\%] $\downarrow$  & ADE [$m$] $\downarrow$ & FDE [$m$] $\downarrow$ & minADE [$m$] $\downarrow$ & minFDE [$m$] $\downarrow$\\
    \midrule 
    \checkmark & $\times$ & 7.44 & 4.44 &4.23 & 17.94 & 3.55 & 14.57\\
    $\times$ & \checkmark & 3.33           & 2.42            & 1.46 & 5.22 & 0.99 & 3.14          \\ \midrule
    \checkmark & \checkmark    & \textbf{2.50}   &\textbf{1.96}    & \textbf{1.42} & \textbf{5.15} & \textbf{0.94} & \textbf{3.01} \\ 
  \bottomrule
  \end{tabular}
\end{table*}


\subsection{Integration of Behavior Prediction}
Another key design choice in {\modelabb} is the integration of a multi-modal behavior predictor within the model architecture as a co-training task. The results shown in \cref{tab: prediction} indicate that the inclusion of the behavior predictor significantly enhances performance in closed-loop simulation tests compared to a baseline approach lacking prediction capabilities. The integration of the predictor contributes to the refinement of the diffusion policy by facilitating more effective learning within the model’s encoder. This interaction strengthens the encoder’s capacity to extract meaningful features, which are critical for downstream tasks. Moreover, the predictor provides behavior priors for individual agents, thereby enabling behavior editing and enhancing the controllability of the diffusion policy.

\begin{table*}[htp]
     \centering
     \caption{Closed-loop Simulation Results with DDIM Sampling}
     \begin{tabular}{@{}l|cccc|cccc@{}}
     \toprule
     \multirow{2}{*}{Step} & \multicolumn{4}{c|}{Unguided Scenario Generation}                                             & \multicolumn{4}{c}{Guided Scenario Generation}\\
                           & \multicolumn{1}{c}{Collision [\%] $\downarrow$} & \multicolumn{1}{c}{Off-road [\%] $\downarrow$} & ADE [$m$] $\downarrow$ & \multicolumn{1}{c|}{Runtime [s] $\downarrow$ }  & \multicolumn{1}{c}{Collision [\%] $\downarrow$} & \multicolumn{1}{c}{Off-road [\%]$\downarrow$ }   & ADE [$m$] $\downarrow$  & \multicolumn{1}{c}{Runtime [s] $\downarrow$} \\ \midrule
     5                     &  2.26   &   1.62 &   1.23 &  $0.16\pm 0.023$       & 1.01    &   1.14  &   1.32    &  $0.55\pm 0.021$    \\
     10                    &  2.39   &   1.72  &  1.25 &  $0.31\pm 0.001$       & 1.06    &   1.35    & 1.35    &  $1.19\pm 0.034$    \\
     25                    &  2.41  &   1.77   &   1.29 &  $0.73\pm 0.002$      & 1.01    &   1.34    & 1.54    &  $2.87\pm 0.078$   \\\midrule 
     DDPM                  &  2.50  &   1.96  &  1.42 &  $1.48\pm 0.005$     & 1.11    &   1.58    & 1.68    &  $6.33\pm0.130$     \\ \bottomrule
     \end{tabular}
     \label{tab:runtime}
\end{table*}

\subsection{Noise Scheduling Design}

{Traditional cosine scheduling tends to favor low noise variance regions during the initial diffusion steps, potentially allowing the model to bypass proper scene representation learning by relying on shortcuts through less noisy inputs. Consequently, when encountering out-of-distribution actions during testing, the model struggles to adapt and instead produces degraded output by directly passing the actions through the final denoising steps. This insight motivated us to develop a logarithmic noise schedule}, presented in \cref{eqn: log_schedule}, which maintains a suitable signal-to-noise ratio (SNR) throughout the diffusion process, ensuring effective action denoising that is dependent on scene context. Additionally, experimental results in \cref{fig: schedule_results} validate the efficacy of our proposed approach, showcasing improved trajectory diffusion performance under the new noise schedule. By adopting the log scheduling strategy, {\modelabb} successfully generates diverse scenarios without compromising generation quality.

The result in \cref{fig: schedule_results} {reveals that the generation quality is satisfactory after just the first denoising step, especially regarding collision rate, off-road rate, ADE, and FDE. Further denoising steps seem to degrade these metrics. One possible explanation is that traffic behavior diffusion is heavily influenced by scene context; hence, the initial denoising step can yield plausible outcomes that adhere to both agent interactions and map constraints. Nevertheless, subsequent denoising steps play a vital role in striking a balance between fidelity and diversity. With increasing numbers of steps, we notice a decline in the minADE and minFDE metrics, suggesting that one of the generated samples draws closer to the ground truth. Average ADE and FDE metrics have increased, highlighting greater sample diversity. This outcome demonstrates that additional denoising steps enable the model to explore a wider range of feasible outcomes while preserving high-quality trajectories. Furthermore, these extra steps are crucial for facilitating control and guidance over agent behaviors, particularly in applications such as traffic scenario generation.}

\subsection{Accelerating Inference} 
Employing Denoising Diffusion Implicit Models (DDIM) \cite{song2020denoising} can significantly accelerate the inference process of the {\modelabb} model while maintaining performance. To evaluate the impact of diffusion steps on simulation and runtime performance, we conduct experiments with varying numbers of diffusion steps in DDIM, and the detailed results are summarized in \cref{tab:runtime}. The runtime of each inference step is measured on a system equipped with an AMD 7950X CPU and an NVIDIA RTX 4090 GPU. The results reveal that the runtime grows linearly with the number of diffusion steps. Notably, employing 5 or 10 diffusion steps achieves a favorable balance between generation quality and real-time performance (with a feasible replan frequency of 1 Hz). Especially when combined with collision cost guidance, we can significantly reduce the collision rate and enhance the quality of the generated behaviors. Specifically, using 5 diffusion steps in DDIM guided sampling yields a notable reduction in collision rates and improves scenario realism without compromising computational efficiency. Moreover, increasing the number of diffusion steps results in diminishing returns for collision avoidance and a negative impact on the log divergence (ADE) metric. This suggests that the performance and efficiency of {\modelabb} can be further enhanced with DDIM sampling to balance computational efficiency and performance while retaining a certain level of controllability.

%% file: sections/conclusions.tex
\section{Conclusions}
We present {\model} (\modelabb), a traffic scenario generation framework that leverages diffusion models to produce realistic, diverse, and controllable multi-agent interactions in closed-loop simulation. Our method demonstrates strong performance in multi-agent behavior simulation and effectively captures complex interaction patterns. A key strength of {\modelabb} lies in its flexible inference-time control, which enables user-guided scenario editing and behavior conditioning to meet specific requirements. Through extensive experiments, we show that {\modelabb} is effective across a broad range of traffic simulation tasks, including safety-critical scenarios.
Several directions remain for future work. A key next step is integrating VBD into AV planning and testing pipelines to enable systematic evaluation and improvement of planning algorithms under generated scenarios, including rare and adversarial cases. Enhancing runtime efficiency will also be critical for real-time deployment in AV simulation and training systems. Additionally, incorporating large language models (LLMs) for context-aware scenario generation and extending the framework to diverse traffic domains, such as highways or urban mixed-autonomy settings, are promising avenues for further development.